\documentclass[journal]{IEEEtran}
\ifCLASSINFOpdf
\else
\fi

\def\eg{\emph{e.g.}}
\def\etal{\emph{et al.\,}}

\usepackage{times}
\usepackage{amsmath}
\usepackage{amssymb}
\usepackage{epsfig}
\usepackage{enumitem}
\usepackage{graphicx}
\usepackage{algorithm}
\usepackage{algorithmic}

\usepackage{lipsum}
\usepackage{multirow}
\usepackage{color}
\usepackage{caption}

\usepackage[
    colorlinks,
    linkcolor=red,
    anchorcolor=red,
    citecolor=green,
]{hyperref}

\begin{document}
\captionsetup[figure]{name={Figure },labelsep=period}
\captionsetup[table]{name={Table },labelsep=period}
\def \vp{\mathrm{\mathbf{v}}}
%
\title{Vanishing Point Guided Natural Image Stitching}
%
%
%


\author{Kai Chen, Jian Yao, Jingmin Tu, Yahui Liu, Yinxuan Li and Li Li\thanks{Kai Chen, Jian Yao, Jingmin Tu, Yinxuan Li and Li Li were with the School of Remote Sensing and Infirmation Engineering, Wuhan University, China. e-mail: chenkai@whu.edu.cn; jian.yao@whu.edu.cn. Yahui Liu was with the University of Trento.}}

\markboth{Journal of \LaTeX\ Class Files,~Vol.~14, No.~8, August~2015}%
{Shell \MakeLowercase{\textit{et al.}}: Bare Demo of IEEEtran.cls for IEEE Journals}
%



\maketitle

\begin{abstract}
	Recently, works on improving the naturalness of stitching images gain more and more extensive attention. Previous methods suffer the failures of severe projective distortion and unnatural rotation, especially when the number of involved images is large or images cover a very wide field of view. In this paper, we propose a novel natural image stitching method, which takes into account the guidance of vanishing points to tackle the mentioned failures. Inspired by a vital observation that mutually orthogonal vanishing points in Manhattan world can provide really useful orientation clues, we design a scheme to effectively estimate prior of image similarity. Given such estimated prior as global similarity constraints, we feed it into a popular mesh deformation framework to achieve impressive natural stitching performances. Compared with other existing methods, including APAP, SPHP, AANAP, and GSP, our method achieves state-of-the-art performance in both quantitative and qualitative experiments on natural image stitching.
\end{abstract}

\vspace{-12pt}

%
\IEEEpeerreviewmaketitle

%
%
%
%

\section{Introduction}
Image stitching is a classical computer vision task that combines multiple images into a panorama with a wider field of view. Methods started to flourish since 2007, when Brown and Lowe~\cite{brown2007automatic} proposed to use SIFT features~\cite{lowe2004distinctive} to fit a global homography model for stitching. Since then, various methods were developed to further improve the stitching performance, including the spatially varying methods~\cite{lin2011smoothly,zaragoza2013projective} with a higher degree of freedom for good alignment accuracy, and the combined-constraints based methods~\cite{li2015dual,lin2017direct,chen2018multiple} for improving stitching robustness. The technique of mesh deformation~\cite{liu2009content} is also adopted to stitching since it has high alignment quality and is highly scalable to some specific stitching purposes, such as stereoscopic stitching~\cite{yan2016reducing,wang2018natural,zhang2015casual}. Unfortunately, these methods usually can not deal with the non-overlapping region well owing to the lack of effective constraints, and as a result, they suffer severe projective distortions. Once number of images gets too large, the distortion would accumulate and propagate among images, leading to unnatural rotation, scaling and stretch.

\begin{figure}[t]
	\centering
	\captionsetup{font={small}}
	\includegraphics[width=0.95\linewidth]{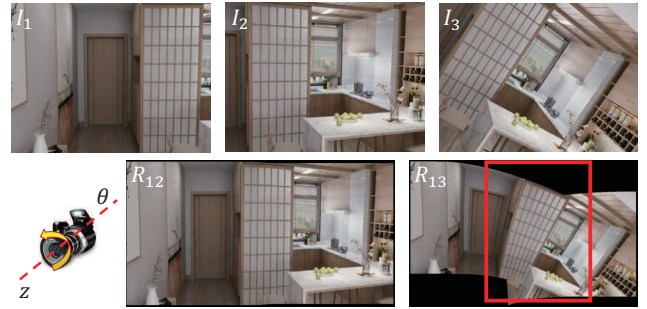}
	\caption{Top: Two pairs of input images $(I_1, I_2)$ and $(I_1, I_3)$. They share the same left image $I_1$, while $I_2$ and $I_3$ differ merely by a 2D rotation $\theta$. Bottom: Two panoramas produced by the SPHP~\cite{chang2014shape}. $R_{12}$ and $R_{13}$ have very different appearances caused by $\theta$.}\vspace{-11pt}
	\label{fig:motivation}
\end{figure}

Natural image stitching methods~\cite{chang2014shape,chen2016natural,li2018quasi,lin2015adaptive} are developed to reduce distortions. Up till now, a widely acceptable fact for natural stitching is to make use of the shape-preserving property~\cite{hartley2003multiple} provided by the similarity transformation. Moreover, the concept of image global similarity prior~(\textit{a.k.a.}, a scale factor $s$ and a 2D rotation angle $\theta$)~\cite{chen2016natural} is proposed to be estimated to further improve the visual quality. Inappropriate rotations between adjacent images often induce obvious unnaturalness in results\footnote{For the presented two-image case, the zero-rotation switch in SPHP was turned on to illustrate the problem. In practical application, the switch has limited effect in improving panorama naturalness for general multiple-image cases. Relevant results will be presented in the following of the paper.}~(as shown in Figure~\ref{fig:motivation}). Thus, the key issue for natural stitching becomes how to properly estimate the image similarity prior. Most existing methods~\cite{chang2014shape,lin2015adaptive,chen2016natural} determine them with matched feature points or line segments. Note that these schemes merely utilize the pairwise correspondences between adjacent images. The absence of a global constraint, which would offer robust guidance for the similarity prior estimation, makes these methods unstable when the overlap between images is small, or the number of the involved images is large. As a result, these methods produce unnatural artifacts.

\begin{figure*}[t]
	\centering
	\captionsetup{font={small}}
	\includegraphics[width=1.0\textwidth]{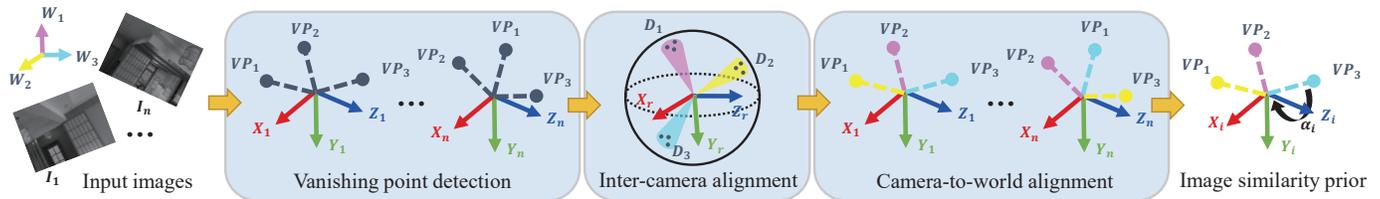}
	\caption{A flowchart for initial rotation estimation $\{\alpha_i\}_{i=1}^N$. VPs are first detected through line segments extraction and clustering. Then, VPs of different images are aligned on a unified sphere surface, on which three dominant VP directions $D$ are estimated. Finally, VPs are associated with the ideal vanishing directions by $D$ to compute $\{\alpha_i\}_{i=1}^N$.}
	\label{fig:flowchart}
\end{figure*}
Considering the mentioned issues, in this paper, we propose to take the vanishing point~(VP) as an effective global constraint, and develop a novel similarity prior estimation method for natural image stitching. We focus on the problem of estimating $\theta$, and exploit the VP guidance by taking its two advantages: (1) utilizing the orientation clues from VPs to estimate the initial 2D rotations for input images; (2) making use of the global consistency of VPs in Manhattan world, by which a novel scheme is proposed to estimate the prior robustly. After that, the determined similarity prior is feed into a mesh deformation framework as global similarity constraints to stitch multiple images into a panorama with a natural look.

In summary, the main contributions of this paper are:
\begin{itemize}
	\item We design a robust scheme to determine the image similarity prior from the VP clues of the scene, based on which a novel natural stitching method named VPG is developed to improve the naturalness of output panoramas significantly.
	\item We provide a degeneration mechanism to make the proposed VPG can be well-used in general scenes. When the scene holds the Manhattan assumption, VPG manages to produce a more natural panorama than other methods. Otherwise, it automatically falls back to a standard stitching scheme. The output still has a relatively natural look that is not affected by wrong VP guidance.
	\item We conducted more analyses upon the proposed VPG algorithm. The results further reveal that VPG has two additional good properties: First, it is not influenced by different reference selections; Second, it is compatible with other high-alignment-accuracy stitching frameworks to achieve a coordination between good naturalness and high alignment accuracy.
\end{itemize}

Abundant quantitative and qualitative experiments both demonstrate that the proposed vanishing point guided method~(VPG) outperforms the state-of-the-art methods, including APAP~\cite{zaragoza2013projective}, SPHP~\cite{chang2014shape}, AANAP~\cite{lin2015adaptive}, and GSP~\cite{chen2016natural}. Intuitive comparisons are available from the project homapage\footnote{http://cvrs.whu.edu.cn/projects/VPGStitching/}.

\section{Related Work}
We briefly review the most related works in three aspects: mesh-deformation based stitching methods, natural image stitching methods, and some existing successful practices in other fields that are associated with the VP guidance.

\subsection{Mesh-deformation based Image Stitching}
Mesh deformation technique~\cite{liu2009content} is adopted to image stitching since its flexibility. It first divides input images into a series of uniform grid meshes, and then estimates deformed mesh vertexes by minimizing an objective function. Various constraints are utilized to build the objective function in order to improve robustness and stitching quality. Early feature point based methods~\cite{zhang2014parallax,zhang2015casual,guo2016joint} detected and matched key-points in the overlapping region, then they achieved alignment by warping matched points to close positions. Li~\etal~\cite{li2015dual} tailored a dual-feature model that considers both key-point and line segment correspondences to perform a robust stitching for low-textured scenes. Lin~\etal~\cite{lin2017direct} were inspired by optical flow estimation and performed alignment by minimizing the overall intensity difference among regularly sampled points. In addition, Xiang~\etal~\cite{xiang2018image} locally regulated image content by penalizing the straightness of line segments, and Lin~\etal~\cite{lin2016seagull} preserved image structures through maintaining the contour shapes. Although these mentioned methods achieve good pairwise alignment accuracy, they are not suited for natural stitching since a lack of valid constraints for the non-overlapping region and the global image content. 

\subsection{Natural Image Stitching}
Most existing methods~\cite{chai2016shape,chang2014shape,chen2016natural,lin2015adaptive,wang2018natural,zhang2016multi}achieved natural stitching by integrating the similarity transformation into a spatially varying homography model. The key issue is how to estimate the global similarity prior for each input image. SPHP~\cite{chang2014shape} determined the similarity transformation through analyzing the pairwise image homography. Then, it enhanced the naturalness via smoothly changing the stitching model from projective to similarity transformation. AANAP~\cite{lin2015adaptive} first computed a bunch of 2D rotations by feature matching, then it empirically selected the similarity transformation with the smallest rotation angle as the optimal one. GSP~\cite{chen2016natural} solved the global similarity prior by feature matching as well as the 3D rotation relationship between adjacent images. VL~\cite{zhang2016multi} built similarity constraints for ortho-photos by taking use of the orientations of line segment clusters. Considering that all these methods lack an effective global guidance, they suffer failures in challenging scenes.

\subsection{Vanishing Point Guidance}
Vanishing point~(VP)~\cite{lu20172} is widely adopted in computer vision tasks since its predominance in two aspects. First, VPs contain strong orientation and geometric clues of a scene which could be useful guidance. Lee~\etal~\cite{lee2009geometric} interpreted the scene structure from VPs extracted from a single image. Lee and Yoon~\cite{lee2015real} recovered the camera orientation with a joint estimation of VPs. Huang~\etal~\cite{huang2014image} exploited VPs in image completion by detecting the planar surfaces and regularity with VPs. Furthermore, VPs were applied in layout estimation~\cite{zou2018layoutnet} to pre-align the image to be level with the floor, and also were utilized to offer geometric context for road detection and recognition~\cite{lee2017vpgnet}. Second, VPs are globally consistent in Manhattan world, and therefore provide an effective global constraint for many optimization-based problems which help to yield stable performance for robot navigation~\cite{lee2009vpass}. Camposeco and Pollefeys~\cite{camposeco2015using} adopted VPs to improve the accuracy of visual-inertial odometry, and Li~\etal~\cite{li2018monocular} leveraged them to build a robust monocular SLAM system. Inspired by these successful applications of VPs, we apply them to natural stitching, and propose a robust similarity prior estimation method making use of the VP guidance, as shown in Figure~\ref{fig:flowchart}.

\section{Our Methodology}\label{sec:methodology}
Given $N$ images $\{I_i\}_{i=1}^N$, we stitch them into a natural panorama using the mesh deformation framework as described previously. Let $V$ be the set of vertexes in the uniform grid mesh that placed on input images, the stitching process is formulated into a mesh deformation problem through finding the optimal warped vertex set $\hat{V}$. Usually, it turns into an optimization problem by minimizing an objective function with the following classical form~\cite{chai2016shape,li2015dual,chen2016natural}:
\begin{equation}\label{eq:classical_mesh_deformation}
E(V)=E_a(V)+E_l(V)+E_g(V),
\end{equation}
where $E_a$, $E_l$ and $E_g$ denote the alignment term, the local shape-preserving term and the global similarity term respectively.

As mentioned, previous methods improve alignment accuracy and conserve image content by regulating $E_a$ and $E_l$, while recent natural stitching methods focus on $E_g$ to produce more natural looking panoramas. $E_g$ is built based on the image similarity priors~(a scale factor $s$ and a 2D rotation $\theta$), but existing methods fail to offer a robust scheme to estimate them. They usually treat each input image separately~\cite{zhang2016multi}, or determine the prior merely with pairwise image correspondence information~\cite{chen2016natural,chang2014shape,lin2015adaptive}, which are not robust enough in practical applications. Therefore, we focus on developing a robust similarity prior estimation method with the guidance of VPs, in which we mainly focus on the estimation for 2D rotations $\{\theta_i\}_{i=1}^N$.

\subsection{Rotation Estimation with VP Guidance}\label{subsec:rotation_estimation}
Let $\mathbf{G}$ be the stitch graph of $N$ input images. $\mathbf{J}$ denotes its edge set, in which each edge $(i, j)$ corresponds to a pair of adjacent images $(I_i, I_j)$. In general, the $\mathbf{G}$ could be manually specified or automatically verified by the probabilistic model~\cite{brown2007automatic}. Let's define $\mathbf{P}=\{\mathbf{P}_{ij}|(i, j)\in\mathbf{J}\}$ as the set of matched feature points. We then apply LSD~\cite{von2010lsd} to detect line segments on $I_i$, and then loosely follow the scheme in~\cite{lee2009geometric} to find three orthogonal VPs $[\vp_1^i, \vp_2^i, \vp_3^i]$ without the intrinsic parameters. As illustrated in Figure~\ref{fig:flowchart}, we subsequently obtain the initial rotation estimation $\{\alpha_i\}_{i=1}^N$ through two steps: (1) \textit{Inter-camera alignment}. VPs from different images are first aligned in the same reference coordinate system, in which we can estimate dominant directions using these roughly aligned VPs. (2) \textit{Camera-world alignment}. VPs are associated with the ideal vanishing directions in Manhattan world to produce initial rotation estimation $\{\alpha_i\}_{i=1}^N$. After that, we design a robust estimation scheme to get optimal rotation result $\{\theta_i\}_{i=1}^N$.

\subsubsection{Inter-Camera Alignment.}We project VPs of different images onto a unified sphere surface in order to achieve alignment. As a preparation, based on the matched point set $\mathbf{P}$, we first estimate the 3D rotation $\mathbf{R}_i$ for each image $I_i$ by bundle adjustment method~\cite{brown2007automatic}. After that, without the loose of generality, we regard $I_r$ as a reference and set $\mathrm{\mathbf{W}}=\mathbf{R}_r$. VPs then are projected as:

\begin{equation}\label{eq:vps_projection}
[\hat{\vp}_1^i, \hat{\vp}_2^i, \hat{\vp}_3^i]=\mathrm{\mathbf{W}}\mathbf{R}_i^{-1}[\frac{\vp_1^i}{\|\vp_1^i\|}, \frac{\vp_2^i}{\|\vp_2^i\|}, \frac{\vp_3^i}{\|\vp_3^i\|}].
\end{equation}
Assuming that the scene satisfies the Manhattan world and considering the possible parallax and estimation noises, these aligned VPs should roughly congregate around three dominant directions $\mathrm{\mathbf{D}}_{3\times3}$. We propose to estimate $\mathrm{\mathbf{D}}_{3\times3}$ by:
\begin{equation}\label{eq:solve_dominant_direction}
\mathrm{\mathbf{D}}=\arg\min_{\mathrm{\mathbf{D}}}\sum_{i=1}^{N}\|\mathrm{\mathbf{D}}_{3\times3}-[\hat{\vp}_1^i, \hat{\vp}_2^i, \hat{\vp}_3^i]\|_{F}^2,
\end{equation}
where $\|\cdot\|_F$ denotes the matrix Frobenius norm. In order to improve the stableness, we decompose $\mathrm{\mathbf{D}}$ into $\mathrm{\mathbf{D}}_0\mathrm{\mathbf{D}}_t$, where $\mathrm{\mathbf{D}}_0$ is the initial dominant direction hypothesis, and $\mathrm{\mathbf{D}}_t$ is a $3\times3$ relative rotation matrix. We collect two roughly orthogonal VPs as the first two directions $\mathbf{d}_1$ and $\mathbf{d}_2$ in $\mathrm{\mathbf{D}}_0$. They produce the third direction by $\mathbf{d}_3=\mathbf{d}_1\times\mathbf{d}_2$. After that, $\mathbf{d}_2$ is revised by $\mathbf{d}_2=\mathbf{d}_1\times\mathbf{d}_3$ to ensure the orthogonality so that we get a complete dominant direction hypothesis $\mathrm{\mathbf{D}}_0=[\mathbf{d}_1,\mathbf{d}_2,\mathbf{d}_3]$. Given $\mathrm{\mathbf{D}}_0$, we can estimate $\mathrm{\mathbf{D}}_t$ by:
\begin{equation}\label{eq:solve_DT}
\mathrm{\mathbf{D}}_t=\arg\min_{\mathrm{\mathbf{D}}_t}\sum_{i=1}^{N}\|\mathrm{\mathbf{D}}_0\mathrm{\mathbf{D}}_t-[\hat{\vp}_1^i, \hat{\vp}_2^i, \hat{\vp}_3^i]\|_{F}^2.
\end{equation}
It is obvious that solving Equation~\ref{eq:solve_DT} is much more stable than directly optimizing Equation~\ref{eq:solve_dominant_direction}. It can be effectively optimized by Gauss-Newton iteration method. After traversing all possible dominant direction hypotheses, we obtain the final result with the minimal residual error as the optimal dominant direction $\hat{\mathrm{\mathbf{D}}}=[\hat{\mathbf{d}}_1,\hat{\mathbf{d}}_2,\hat{\mathbf{d}}_3]$.

\subsubsection{Camera-to-World Alignment.}Let the three global VPs associated with three dominant directions in Manhattan world be $[\vp_1^{\mathrm{w}}, \vp_2^{\mathrm{w}}, \vp_3^{\mathrm{w}}]=[[1,0,0]^{\mathsf{T}}, [0,1,0]^{\mathsf{T}}, [0,0,1]^{\mathsf{T}}]$. We assume that people rarely twist the camera severely relative to the horizon when capturing a picture, which can be a relative loose assumption than the one in \cite{brown2007automatic} and \cite{chen2016natural}. Hence, we associate $\hat{\mathrm{\mathbf{D}}}$ with $[\vp_1^{\mathrm{w}}, \vp_2^{\mathrm{w}}, \vp_3^{\mathrm{w}}]$ by $\mathrm{\mathbf{M}}=[\mathbf{m}_1,\mathbf{m}_2,\mathbf{m}_3]$, which is determined by
\begin{equation}
\mathbf{m}_i=\arg\min_{\vp_j^{\mathrm{w}}}\|\hat{\mathbf{d}}_i-\vp_j^{\mathrm{w}}\|^2, i,j=1,2,3.
\end{equation}

Meanwhile, we rearrange $[\hat{\vp}_1^i, \hat{\vp}_2^i, \hat{\vp}_3^i]$ for $I_i$ in order to make them correspond to $\hat{\mathrm{\mathbf{D}}}$, that is, $\hat{\vp}_{k}^i\leftrightarrow\hat{\mathbf{d}}_{k},k=1,2,3$. Then, the transformation from the $i$-th camera to the global Manhattan world is formulated as:
\begin{equation}
\mathbf{R}_i^{\mathrm{w}}=\mathrm{\mathbf{M}}[\hat{\vp}_1^i, \hat{\vp}_2^i, \hat{\vp}_3^i]^{-1},
\end{equation}
where $\mathbf{R}_i^{\mathrm{w}}$ is a $3\times3$ rotation matrix. We decompose it to get $\alpha_i$, which is a 2D rotation angle with respect to $z$ axis.
\begin{figure}[t]
	\centering
	\captionsetup{font={small}}
	\includegraphics[width=1.0\columnwidth]{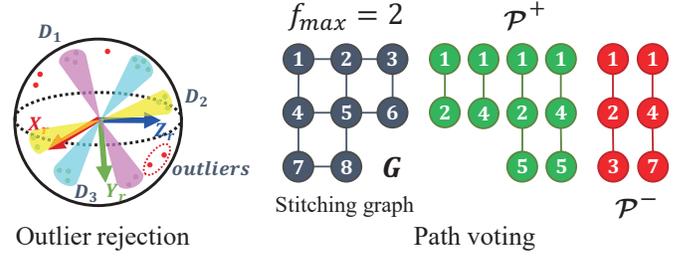}
	\caption{An illustration of two strategies adopted for robustly estimating $\{\theta_i\}_{i=1}^N$. Left: Given the VPs alignment result and dominant directions, images whose VPs has large residuals are marked as outliers. Right: Initial rotations are propagated along the stitching graph, and are weighted using a path voting scheme.}\vspace{-15pt}
	\label{fig:robust_estimation}
\end{figure}

\subsubsection{Robust Estimation.}Instead of directly taking $\alpha_i$ as $\theta_i$, we propose to estimate $\{\theta_i\}_{i=1}^N$ through minimizing an objective function with regarding $\alpha_i$ as the data term. In addition, we consider the relative rotation between adjacent images as the smoothness term. Specifically, for image pair $(I_i,I_j)$, the relative 2D rotation $\beta_{i,j}$ is obtained by decomposing $\mathbf{R}_j\mathbf{R}_i^{-1}$. We represent $\alpha_i$ by a unit 2D vector $(\phi_i,\omega_i)^{\mathsf{T}}$, and the rotation $\{\theta_i=(u_i,v_i)^{\mathsf{T}}\}_{i=1}^N$ are obtained by minimizing:

\begin{equation}\label{eq:estimate_theta}
\sum_{i=1}^N\|\left[\begin{matrix}
u_i\\v_i
\end{matrix}\right]-\left[\begin{matrix}
\phi_i\\\omega_i
\end{matrix}\right]\|^2+\lambda\sum_{(i,j)\in\mathbf{J}}\|\mathbf{R}(\beta_{i,j})\left[\begin{matrix}
u_i\\v_i
\end{matrix}\right]-\left[\begin{matrix}
u_j\\v_j
\end{matrix}\right]\|^2,
\end{equation}
where $\mathbf{R}(\beta_{i,j})$ denotes the $2D$ rotation matrix specified by $\beta_{i,j}$, and $\lambda=10.0$ is a balance weight. In Equation~\ref{eq:estimate_theta}, we consider each $\alpha_i$ equally, which is easily affected by possible noises existing in $\{\alpha_i\}_{i=1}^N$. Therefore, as shown in Figure~\ref{fig:robust_estimation}, two strategies are further designed to deal with this problem.

\textit{\textbf{Outlier rejection}}. Previously, we have roughly aligned VPs of different images and have obtained their three dominant directions $\hat{\mathrm{\mathbf{D}}}$. Starting from the global consistency of VPs in Manhattan world, we compute the residual difference $e_i$ for $I_i$ according to its $[\hat{\vp}_1^i, \hat{\vp}_2^i, \hat{\vp}_3^i]$ and $\hat{\mathrm{\mathbf{D}}}$. The image whose $e_i$ is larger than a threshold $\tau$ is marked as an outlier. We only compute $\alpha_i$ for inliers, and adopt them as the data term in Eq.~\ref{eq:estimate_theta} to estimate the image similarity prior.

\textit{\textbf{Path voting}}. The scheme of outlier rejection actually is a hard constraint. It has to face a dilemma that a too small $\tau$ may wrongly filter out many images. Otherwise, it may introduce some fallacious $\alpha_i$. A path voting scheme acts as a soft constraint to cope with this dilemma.

As shown in Figure~\ref{fig:robust_estimation}, given the stitching graph $\mathbf{G}$ with $N$ images, we collect all valid paths $\mathcal{P}_i=\{p_i^k\}_{k=1}^{c_i}$ within a maximal length $f_{max}$ for each inlier image $I_i$, where $c_i$ denotes the total number of valid paths for $I_i$. $p_i^k$ starts from $I_i$ and ends at its neighboring inlier images. A path is valid only when it does not pass through any outlier image. We then divide $\mathcal{P}_i$ into two parts: the supporting set $\mathcal{P}_i^{+}$ and the opposing set $\mathcal{P}_i^{-}$. Such a division is achieved based on the relative rotation angle $\beta_{i,j}$ between adjacent images and by judging each $p_i^k$ whether it supports the estimation result $\alpha_i$ for image $I_i$ or not. For $p_i^k\in\mathcal{P}_i^{+}$, we directly take its path length $L(p_i^k)$ as the supporting length $S(p_i^k)$. Otherwise, its opposing length $O(p_i^k)$ is calculated as $f_{max}+1-L(p_i^k)$. Then, we weight the corresponding data term with $\psi_i$, which is defined as:
\begin{equation}\label{eq:path_voting_weight}
\psi_i = \sigma(\frac{\sum_{p\in\mathcal{P}_i^{+}}S(p)}{\sum_{p\in\mathcal{P}_i^{+}}S(p)+\sum_{p\in\mathcal{P}_i^{-}}O(p)}),
\end{equation}
where $\sigma(\cdot)$ is a sigmoid-form non-linear kernel function. 
After the above two schemes, the data term in Eq.~\ref{eq:estimate_theta} is developed into:
\begin{equation}\label{eq:robust_estimate_theta}
\sum_{I_i\in\Psi}\psi_i\|\left[\begin{matrix}
u_i\\v_i
\end{matrix}\right]-\left[\begin{matrix}
\phi_i\\\omega_i
\end{matrix}\right]\|^2,
\end{equation}
where $\Psi$ denotes inlier images. $\{\theta_i\}_{i=1}^N$ are obtained by minimizing Equation~\ref{eq:estimate_theta}.

\subsection{Stitching by Mesh Deformation}\label{subsec:mesh_deformation}
After collecting $\{\theta_i\}_{i=1}^{N}$, we need to determine the scale factor $\{s_i\}_{i=1}^{N}$ to build a complete global similarity constraint. Chen and Chuang~\cite{chen2016natural} propose to estimate $s_i$ by the ratio of focal length, yet this scheme relies too much on camera intrinsic matrix. In contrast, we resort to the matched point set $\mathbf{P}=\{\mathbf{P}_{ij}|(i, j)\in\mathbf{J}\}$ to determine $s_i$. For a pair of adjacent images $(I_i,I_j)$, we first estimate their relative scale $\eta^{ij}$ by:
\begin{equation}
\eta^{ij}=\frac{c(h_i)}{c(h_j)},
\end{equation}
where $h_i$ and $h_j$ are two convex hulls that are determined by $\mathbf{P}_{ij}$, and $c(\cdot)$ returns the perimeter of a convex hull. After that, we estimate the absolute scale $s_i$ for $I_i$ by solving:\vspace{-3pt}
\begin{equation}
\arg\min_{s}\sum_{(i,j)\in\mathbf{J}}\|\eta^{ij}s_j-s_i\|^2, s.t. \sum_{i=1}^{N}s_i=N.\vspace{-3pt}
\end{equation}
It is a quadratic constrained minimization problem, and can be efficiently solved by any linear system. With $\{\theta_i\}_{i=1}^N$ and $\{s_i\}_{i=1}^{N}$, we take the deformation objective function~(Eq.~\ref{eq:classical_mesh_deformation}) from~\cite{chen2016natural} as our baseline, but boost its $E_g$ with our improved image similarity prior. The final stitched panorama eventually is generated by texture mapping technique.

\section{Experiments and Results}\label{sec:experiments}
In this section, we compared our proposed VPG with four state-of-the-art methods: APAP~\cite{zaragoza2013projective}, SPHP~\cite{chang2014shape}, AANAP~\cite{lin2015adaptive}, and GSP~\cite{chen2016natural}. Besides the widely used qualitative comparison manner, two metrics were designed based on the collected synthetic image sets to quantitatively asses the panorama naturalness produced by different methods. It is encouraged to browse the website, \url{http://cvrs.whu.edu.cn/projects/VPGStitching/}, in which more vivid results are provided for clear observation and comparison.\vspace{-7pt}

\begin{figure*}[t]
	\centering
	\captionsetup{font={small}}
	\includegraphics[width=1.0\linewidth]{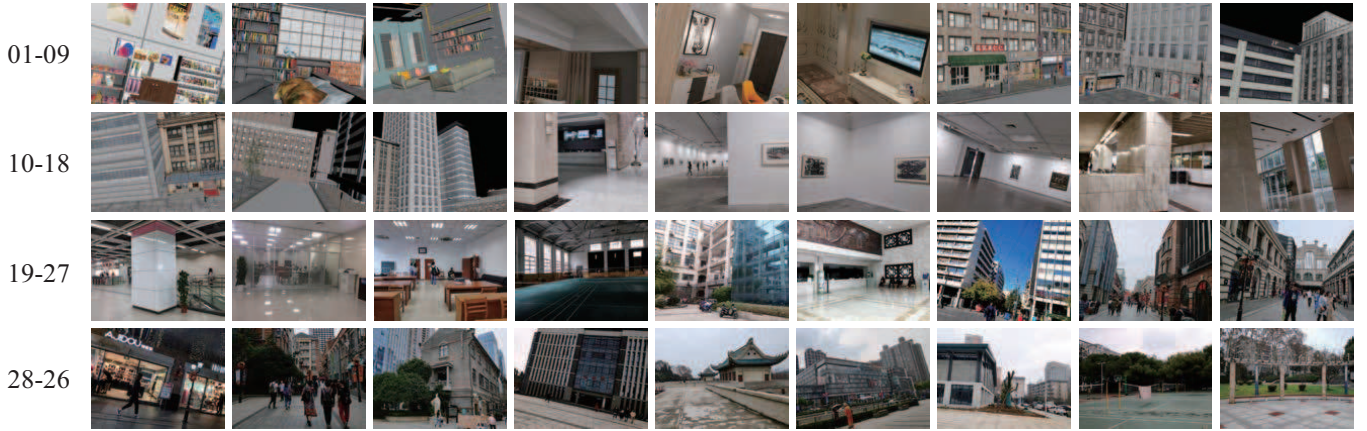}
	\caption{An overview of the VPG dataset. It consists of $12$ synthetic sets~(01-12) and $24$ real sets~(13-36). 01-06 and 13-24 are indoor cases. 07-12 and 25-36 are outdoor street-view cases.}
	\label{fig:VPG_Dataset}
\end{figure*}

\subsection{VPG Dataset}\label{subsec:datasets}
$36$ sets of images were collected to form the VPG dataset. As shown in Figure~\ref{fig:VPG_Dataset}, it includes $12$ sets of synthetic images and $24$ sets of real images. All synthetic images were generated through 3Ds Max rendering\footnote{https://www.autodesk.com/} hence the associated camera parameters are known. All real images were captured by ourselves with a mobile phone. The VPG dataset contains both indoor scenes and outdoor street-view scenes. Specifically, $12$ synthetic sets are composed of $6$ indoor cases and $6$ street-view cases. Similarly, $24$ real sets consist of $12$ indoor scenes and $12$ street-view scenes. More details can be found in Figure~\ref{fig:VPG_Dataset}. It is noteworthy that all images were carefully collected to ensure that they satisfy the Manhattan assumption. The number of images involving in stitching in each set ranges from $5$ to $72$.\vspace{-7pt}

\subsection{Quantitative Metrics}\label{subsec:metrics}
Before specific experiments, two quantitative metrics first are introduced for the assessment of panorama naturalness. Note that comprehensively evaluating the panorama quality is still an open research problem~\cite{cheung2017content,ling2018no}, and it is not the main concern of this paper either. Therefore, we simply start from the observation that a panorama produced by stitching methods usually suffers two kinds of unnaturalness: local projective distortion, and global unnatural rotation. Accordingly, we put forward two indexes for quantitative evaluation: Local Distortion~(LD) and Global Direction InConsistency~(GDIC).

\begin{figure}[t]
	\centering
	\captionsetup{font={small}}
	\includegraphics[width=1.0\columnwidth]{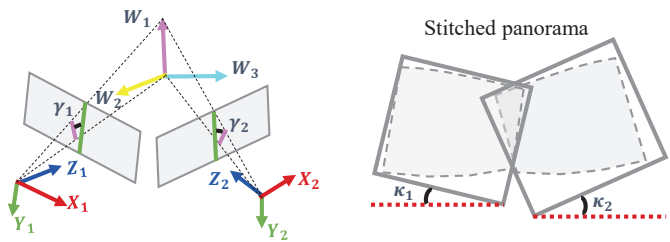}
	\caption{An illustration of the proposed GDIC metric. With the assist of external parameters, we approximate the direction of image content by a 2D angle $\gamma$ with respect to the absolute vertical direction. Accordingly, we approximate the direction of stitched image content by $\kappa$, the orientation of the image bounding rectangle. The GDIC then measures the average difference between these two types of directions.}
	\label{fig:GDIC}
\end{figure}
\vspace{3pt}
\subsubsection{\textit{\textbf{LD-index}}} LD is used to evaluate the local projective distortion. 
Let $Q$ be a quad with four vertexes, the local homography $H$ can be computed from its original and deformed coordinates. We first use a similar way as in~\cite{chang2014shape} to measure the local area change at image position $(x,y)$:
\begin{equation}
\mathcal{M}(H,(x,y))=\text{det}\; J_H(x,y),
\end{equation}
where $J_H$ is the Jacobian of $H$. We then calculate the mean $\mu_Q$ and the standard deviation $\sigma_Q$ of $\mathcal{M}$ from all pixels within $Q$, and take the coefficient of variation~(c.v.) $\sigma_Q/\mu_Q$ as the measurement of projective distortion for $Q$. Subsequently, for $I_i$, we compute the c.v. for all quads that located on the non-overlapping region, and take their average value as the measurement $\mathcal{D}_{i}$ for $I_i$. Finally, we define the LD index as:
\begin{equation}
\text{LD}=\max(\mathcal{D}_{1}, \mathcal{D}_{2},...,\mathcal{D}_{N}).
\end{equation}
\subsubsection{\textit{\textbf{GDIC-index}}} GDIC aims to measure the global unnatural rotation. Assuming that the external parameters are known for $I_i$, as shown in Figure~\ref{fig:GDIC}, we compute a rotation $\gamma_i$ for $I_i$ on the image plane. It is a 2D angle between the camera $y$-axis and the absolute vertical direction. $\gamma_i$ approximates the direction of image content in the world coordinate system. Accordingly, on the output panorama,  we estimate the bounding rectangle with the minimal area for deformed vertexes of $I_i$, and take the rectangle orientation $\kappa_i$ as the direction of $I_i$ after stitching. We think that the relative direction of image content should be preserved as much as possible if images are stitched with a natural look, and the GDIC then is defined as:
\begin{equation}\label{eq:gdic}
\text{GDIC}=\frac{\sum_{i=1,i\ne r}^{N}|(\kappa_i-\kappa_r)-(\gamma_i-\gamma_r)|}{N-1},
\end{equation}
where $r$ denotes the selected reference for relative direction computation. $I_r$ is fixed when computing GDIC for different methods. Since the proposed GDIC requires the external parameters, it is available only for synthetic image data in our experiments.

\subsection{Comparison with APAP~\cite{zaragoza2013projective} and SPHP~\cite{chang2014shape}}\label{subsec:experiment_apap}
\begin{figure*}[t]
	\centering
	\captionsetup{font={small}}
	\includegraphics[width=0.9\linewidth]{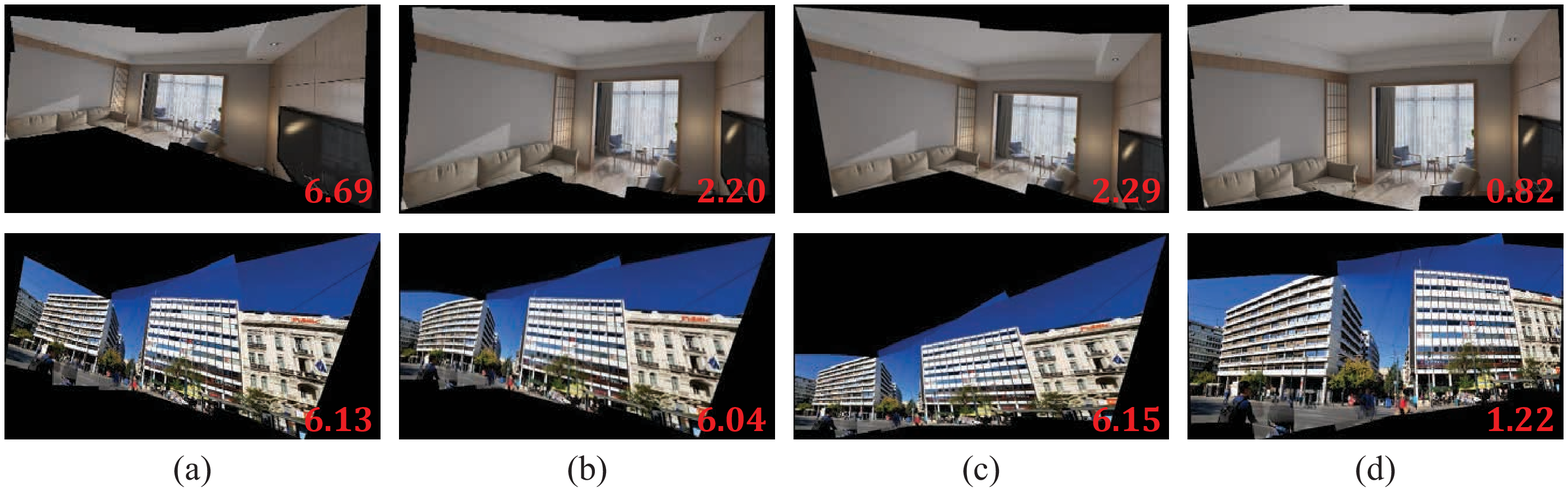}
	\captionof{figure}{Qualitative comparisons with APAP~\cite{zaragoza2013projective} and SPHP~\cite{chang2014shape}. (a) results from APAP. (b) results from SPHP with the zero-rotation switch on. (c) results from SPHP* with the zero-rotation switch off. (d) results from our VPG. The corresponding LD values are given in red text.}
	\label{fig:qualitative_apap}
\end{figure*}

We first compared VPG with two early state-of-the-art algorithms: APAP~\cite{zaragoza2013projective} and SPHP~\cite{chang2014shape}. They were tested using the source code provided by the authors. Since APAP and SPHP have limitations on the field of view, we had to reduce the number of involved images from tens to $3$-$9$ during our experiments. Table~\ref{table:comparison_apap} reports the quantitative results on synthetic sets $04$-$09$. VPG outperforms other two competing methods in both LD and GDIC metrics. Figure~\ref{fig:qualitative_apap} further provides qualitative comparisons on one synthetic set~($04$) and one real set~($29$). As we can see, panoramas produced by APAP and SPHP exhibit severe projective distortions and unnatural rotations. Adjusting the zero-rotation switch for SPHP has very limited effects in improving naturalness. In contrast, the proposed VPG produces panoramas with apparently higher visual quality. The qualitative comparison also is consistent with the quantitative evaluation.

\begin{figure}[t]
	\captionsetup{font={small}}
	\begin{minipage}[t]{1.0\linewidth}
		\renewcommand{\tabcolsep}{1.0pt}
		\renewcommand\arraystretch{1.1}
		\captionof{table}{Quantitative comparisons with APAP~\cite{zaragoza2013projective} and SPHP~\cite{chang2014shape} on synthetic image set $04$-$09$. SPHP* denotes the zero-rotation switch is turned off during stitching.}
		\label{table:comparison_apap}
		\centerline{\begin{tabular}{|c|c|c|c|c|c|c|c|}
				\hline
				Metrics & Methods & 04 & 05 & 06 & 07 & 08 & 09\\\cline{1-8}
				\multirow{4}{*}{LD~($\times10^{-2}$)\hspace{-7pt}$\quad\downarrow$} & APAP & 6.69 & 8.36 & 6.70 & 4.01 & 6.21 & 5.55\\
				{} & SPHP & 2.20 & 3.93 & 7.32 & 2.66 & 3.07 & 2.32\\
				{} & SPHP* & 2.29 & 2.08 & 3.17 & 2.81 & 2.29 & 2.16\\
				{} & VPG & \textbf{0.82} & \textbf{1.35} & \textbf{1.11} & \textbf{1.08} & \textbf{0.19} & \textbf{1.15}\\\hline
				\multirow{4}{*}{GDIC~(deg)\hspace{-7pt}$\quad\downarrow$} & APAP & 18.94 & 22.76 & 11.44 & 11.48 & 4.76 & 7.02\\
				{} & SPHP & 11.21 & 20.16 & 5.32 & 4.92 & 8.06 & 13.27\\
				{} & SPHP* & 8.75 & 3.94 & 6.58 & 8.52 & 6.30 & 12.70\\
				{} & VPG & \textbf{1.21} & \textbf{0.81} & \textbf{0.95} & \textbf{0.95} & \textbf{1.16} & \textbf{2.13}\\\hline
		\end{tabular}}
	\end{minipage}
\end{figure}

\begin{table*}[tbh]
	\captionsetup{font={small}}
	\begin{minipage}[t]{1.0\textwidth}
		\centering
		\renewcommand{\tabcolsep}{5.5pt}
		\renewcommand\arraystretch{1.1}
		\caption{Quantitative comparisons with AANAP~\cite{lin2015adaptive} and GSP~\cite{chen2016natural} on $12$ sets of synthetic images. 01-06 are indoor scenes, and 07-12 are outdoor street-view scenes.}
		\label{table:comparison_gsp}
		\centering
		\normalsize
		\begin{tabular}{|c|c|c|c|c|c|c|c|c|c|c|c|c|c|}
			\hline
			Metrics & Methods & 01 & 02 & 03 & 04 & 05 & 06 &07 & 08 & 09 & 10 & 11 & 12\\\cline{1-14}
			\multirow{4}{*}{LD~($\times10^{-2}$)\hspace{-7pt}$\quad\downarrow$} & AANAP & \textbf{1.21} & 1.56 & \textbf{1.20} & 2.94 & \textbf{1.47} & \textbf{1.10} & 0.94 & 5.17 & 5.95 & 1.50 & 1.14 & 0.89\\
			{} & GSP-2D & 1.48 & 1.73 & 1.82 & 2.19 & 1.85 & 1.28 & 1.00 & 3.00 & 2.68 & 2.04 & \textbf{1.01} & 0.74\\
			{} & GSP-3D & 1.41 & 1.60 & 1.59 & \textbf{1.81} & 1.84 & 1.43 & 1.05 & 2.06 & 2.69 & 1.35 & 1.11 & \textbf{0.67}\\
			{} & VPG & 1.44 & \textbf{1.45} & 1.57 & 2.06 & 1.68 & 1.53 & \textbf{0.93} & \textbf{2.69} & \textbf{2.50} & \textbf{1.27} & 1.19 & 0.73\\\hline
			\multirow{4}{*}{GDIC~(deg)\hspace{-7pt}$\quad\downarrow$} & AANAP & 2.26 & 1.06 & 2.33 & 8.72 & 7.24 & 8.60 & 2.75 & 9.49 & 25.06 & 4.37 & 2.59 & 3.25\\
			{} & GSP-2D & 6.06 & 2.07 & 1.66 & 2.77 & 6.20 & 1.63 & 2.50 & 2.70 & 4.26 & 1.23 & 2.22 & 1.76\\
			{} & GSP-3D & 4.54 & 2.78 & 3.41 & 1.05 & 1.83 & 1.05 & 3.73 & 2.05 & 0.86 & \textbf{1.11} & 1.70 & 1.18\\
			{} & VPG & \textbf{0.73} & \textbf{0.49} & \textbf{0.77} & \textbf{0.49} & \textbf{0.55} & \textbf{0.74}& \textbf{0.72} & \textbf{0.50} & \textbf{0.52} & 1.41 & \textbf{0.81} & \textbf{0.63}\\\hline
		\end{tabular}
	\end{minipage}
	\vfill
	\begin{minipage}{1.0\textwidth}
		\centering
		\includegraphics[width=1.0\linewidth]{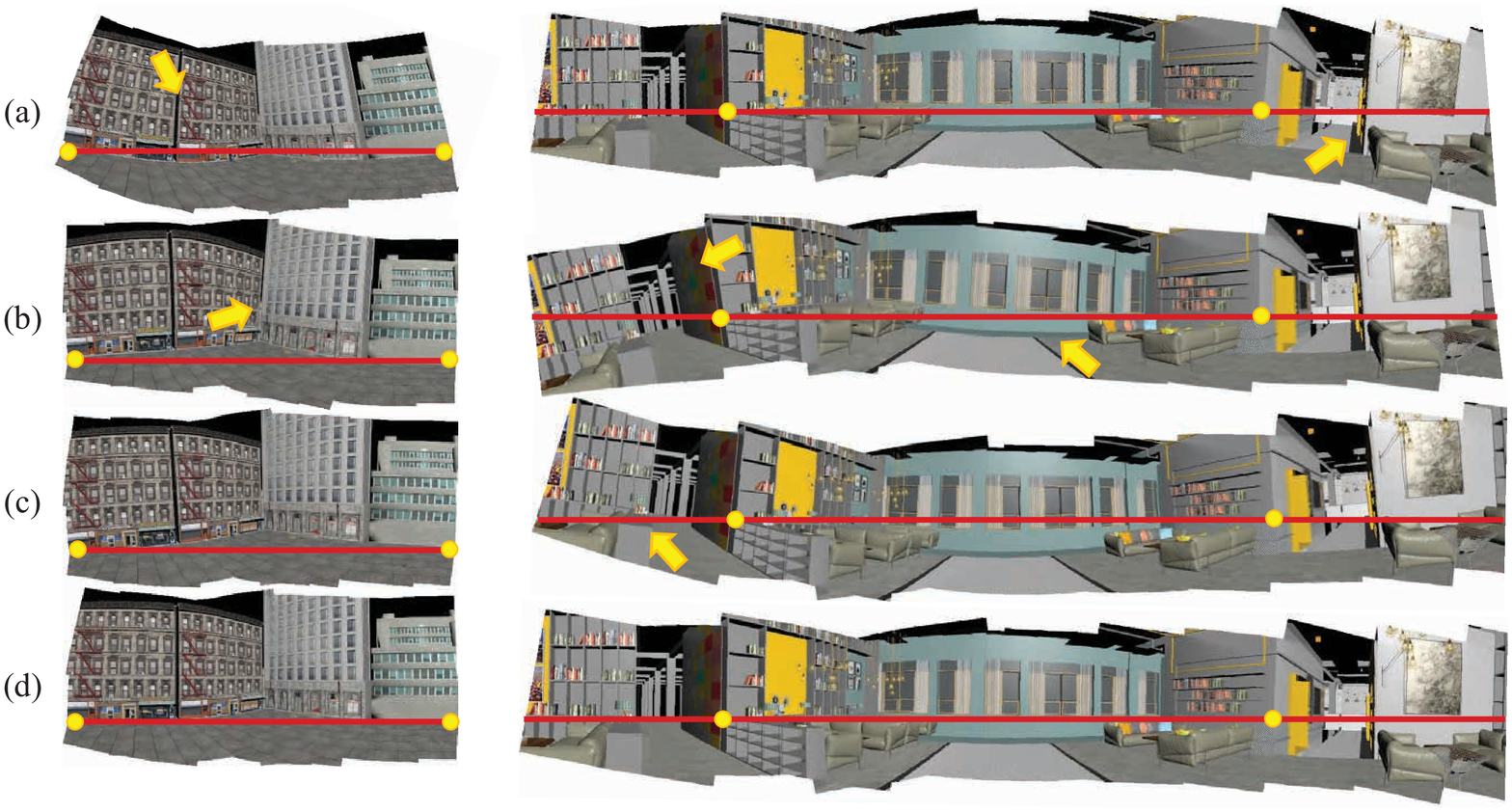}
		\captionof{figure}{Qualitative comparisons with AANAP~\cite{lin2015adaptive} and GSP~\cite{chen2016natural} on synthetic set $03$ and $08$. (a) results from AANAP. (b) results from GSP-2D. (c) results from GSP-3D. (d) results from the proposed VPG. The yellow points indicating the anchor directions that have been aligned to the red horizontal lines for better visual comparison. The yellow arrows hightlight the unnatural artifacts. The same marks are adopted in the following figures.}
		\label{fig:qualitative_aanap_synthetic}
	\end{minipage}
\end{table*}

\subsection{Comparison with AANAP~\cite{lin2015adaptive} and GSP~\cite{chen2016natural}}\label{subsec:experiment_gsp}
We compared the proposed VPG with two recent state-of-the-art natural stitching methods: AANAP~\cite{lin2015adaptive} and GSP~\cite{chen2016natural}. GSP was tested using the source code provided by authors and both the 2D solution and the 3D solution were tested. AANAP was tested using our own re-implementation. Table~\ref{table:comparison_gsp} offers quantitative comparisons on synthetic image sets $01-12$. The proposed VPG has a comparable performance with AANAP and GSP in LD metric since they all share a similar mesh deformation framework to reduce projective distortion. Moreover, in most cases, VPG steadily produces the smallest GDIC values, which means the best global natural look among four methods of comparison. Figure~\ref{fig:qualitative_aanap_synthetic} and Figure~\ref{fig:qualitative_aanap_real} further present typical qualitative comparisons on synthetic and real images respectively, from which the superiority of proposed VPG can be observed intuitively. AANAP empirically estimates the image rotation with the smallest angle, GSP-2D assumes the zero rotation for images, and GSP-3D determines the rotation with pairwise 3D rotation relationships. Their results exhibit obvious unnaturalness since the lack of effective global constraints. In contrast, the proposed VPG manages to improve the panorama naturalness significantly.


\begin{table*}[t]
	\captionsetup{font={small}}
	\begin{minipage}[t]{1.0\textwidth}
		\centering
		\includegraphics[width=1.0\linewidth]{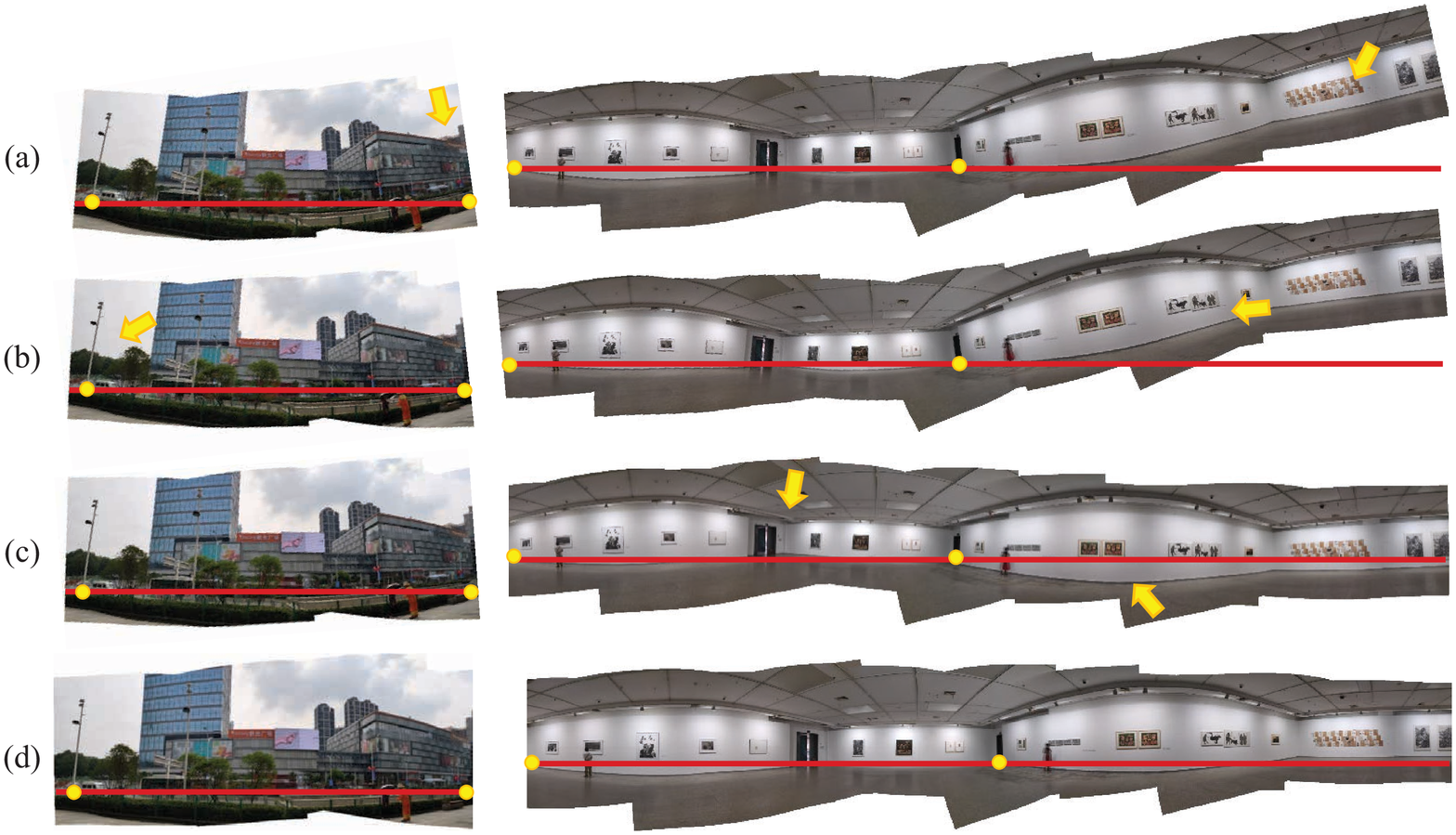}
		\captionof{figure}{Qualitative comparisons with AANAP~\cite{lin2015adaptive} and GSP~\cite{chen2016natural} on real image set $16$ and $33$. (a) results from AANAP. (b) results from GSP-2D. (c) results from GSP-3D. (d) results from our VPG.}
		\label{fig:qualitative_aanap_real}
	\end{minipage}
	\vfill
	\vspace{5pt}
	\begin{minipage}[t]{1.0\textwidth}
		\centering
		\renewcommand{\tabcolsep}{5.5pt}
		\renewcommand\arraystretch{1.1}
		\caption{Quantitative validation of the proposed robust estimation scheme for the image similarity prior. The GDIC values of $12$ synthetic image sets are reported.}
		\label{table:validation}
		\centering
		\normalsize
		\begin{tabular}{|c|c|c|c|c|c|c|c|c|c|c|c|c|c|}
			\hline
			Noises & Methods & 01 & 02 & 03 & 04 & 05 & 06 &07 & 08 & 09 & 10 & 11 & 12\\\cline{1-14}
			\multirow{2}{*}{0\% noises} & VPG~(w/o robust scheme) & \textbf{0.73} & \textbf{0.49} & \textbf{0.77} & 0.82 & 1.01 & \textbf{0.74} & \textbf{0.72} & 0.65 & 0.96 & 2.41 & \textbf{0.81} & \textbf{0.63}\\
			{} & VPG~(w/ robust scheme) & \textbf{0.73} & \textbf{0.49} & \textbf{0.77} & \textbf{0.49} & \textbf{0.55} & \textbf{0.74} & \textbf{0.72} & \textbf{0.50} & \textbf{0.52} & \textbf{1.41} & \textbf{0.81} & \textbf{0.63}\\\hline
			\multirow{2}{*}{10\% noises} & VPG~(w/o robust scheme) & 2.97 & 0.44 & 1.28 & 1.74 & 0.90 & 0.80 & 0.65 & 1.65 & 0.81 & 2.30 & 1.01 & 1.79\\
			{} & VPG~(w/ robust scheme) & \textbf{0.53} & \textbf{0.44} & \textbf{0.64} & \textbf{0.84} & \textbf{0.59} & \textbf{0.65} & \textbf{0.65} & \textbf{0.54} & \textbf{0.38} & \textbf{1.37} & \textbf{0.94} & \textbf{1.53}\\\hline
			\multirow{2}{*}{20\% noises} & VPG~(w/o robust scheme) & 4.74 & 2.57 & 4.47 & 2.29 & 2.47 & 2.67 & 1.67 & 2.21 & 3.19 & 8.55 & 1.65 & 1.13\\
			{} & VPG~(w/ robust scheme) & \textbf{0.56} & \textbf{0.42} & \textbf{0.63} & \textbf{0.47} & \textbf{0.57} & \textbf{0.63} & \textbf{0.74} & \textbf{0.71} & \textbf{0.56} & \textbf{3.75} & \textbf{0.94} & \textbf{0.85}\\\hline
		\end{tabular}
	\end{minipage}
\end{table*}


\subsection{User Study}\label{subsec:experiment_userstudy}
\begin{figure}[t]
	\centering
	\captionsetup{font={small}}
	\includegraphics[width=0.7\linewidth]{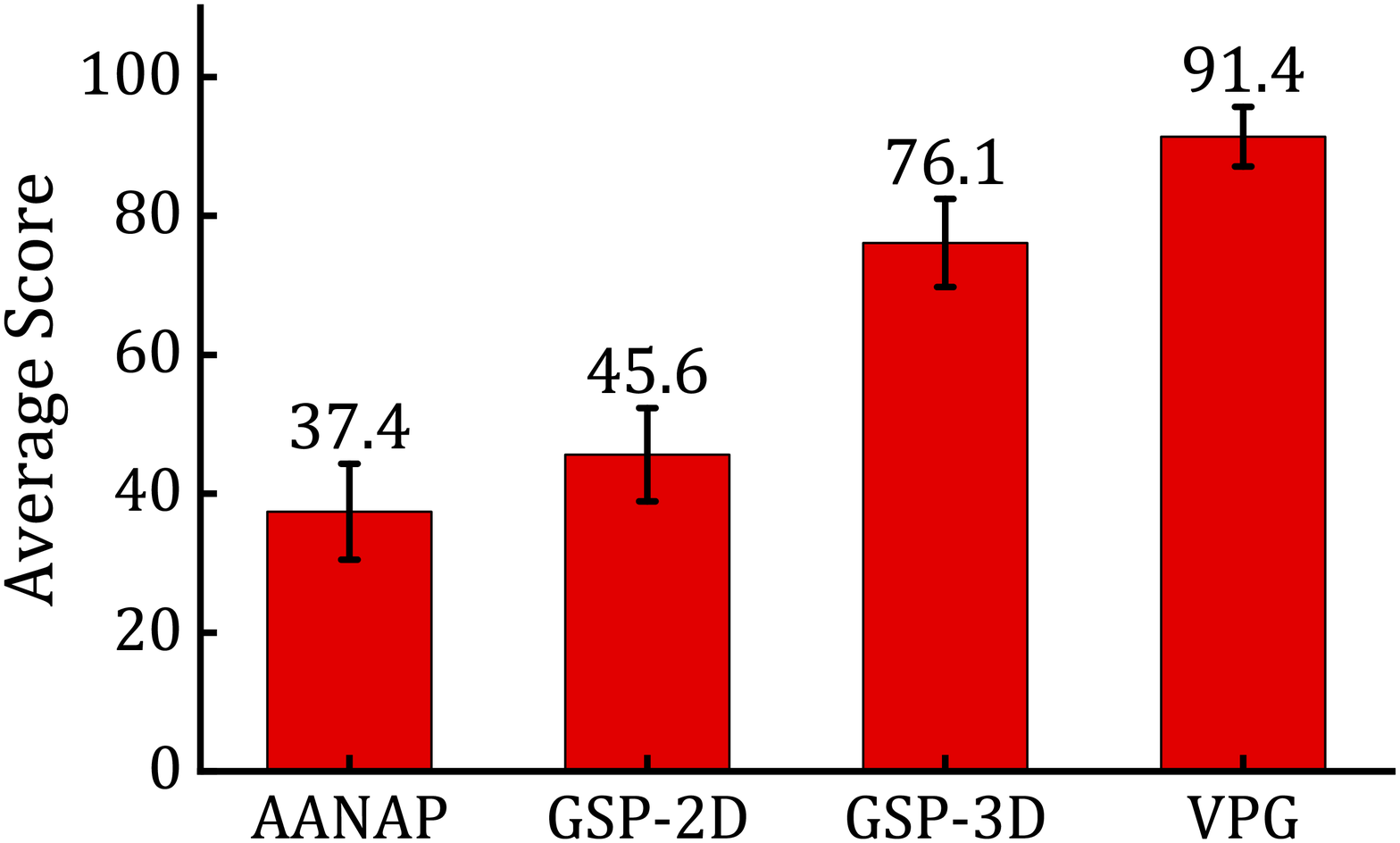}
	\captionof{figure}{User study result. The average scores of $4$ methods from $20$ participants.}
	\label{fig:user_study}
\end{figure}

Since naturalness is a subjective feeling, we further conducted a user study to investigate whether the proposed VPG is preferred by users. In practice, we invited 20 participants, including $10$ researches/students with computer vision/graphics backgrounds and remaining 10 volunteers outside this community. We randomly selected 20 groups of
stitching results in different scenes~(\eg, indoor and street-view) for the user study. There were $4$ unannotated panoramas in each group that were produced by $4$ methods: AANAP, GSP-2D, GSP-3D, and our VPG. Panoramas were shown on a screen in sequence, and the user was allowed to switch images back and forth for a convenient comparison. Then, each participant ranked four results in each group, and assigned each panorama with the corresponding score~(from rank 1 to rank 4, scores varies from 5 to 2). Figure~\ref{fig:user_study} shows the user study results. The VPG is substantially preferred. In addition, it indicates to some extent that the adopted two metrics, LD and GDIC, are consistent with the user's subjective evaluation.

\subsection{Validation of the Robust Estimation Scheme}\label{subsec:experiment_validation}
VPG determines the image similarity prior using a two-step robust estimation scheme. In this section, we hope to quantitatively valid the effectiveness of this step. Using the VPG synthetic image set $01$-$12$ and given the VP extraction results from different images, we manually added random noises to the VP coordinates before adopting them to extract VPs guidance. Then, we compared the associated GDIC values produced by VPG without the robust estimation scheme and by VPG with the robust estimation scheme. Table~\ref{table:validation} reports the comparison results when gradually increasing the noise ratio from $0\%$ to $20\%$. As we can see, metric values from VPG~(w/o) increases significantly as the noises increase while values from VPG~(w/) keep relatively stable in most cases. It demonstrates that the proposed robust estimation scheme effectively ensures the robustness of VPG and maintains the consistently high naturalness of the output panoramas.

\section{More Analysis}\label{sec:analysis}
We further analyzed the performance of VPG in four main aspects: (1) the adaptability for general scenes where the Manhattan assumption may not hold; (2) the stability for reference selection; (3) the scalability for higher alignment accuracy; (4) the time efficiency. 

\subsection{Adaptability for General Scenes}\label{subsec:analysis_generalscene}
We explored the possibility that VPG adaptively falls back to a regular stitching scheme without using the VP clues when the scene disobeys the required Manhattan assumption. Since we have projected the VPs from different images on a unified sphere surface in previous Sections, we think these roughly aligned VPs can reflect the regularity of the scene.

Given a set of aligned VPs, we define its associated VP divergence as follows:
\begin{equation}\label{eq:vp_divergence}
\varepsilon=\frac{1}{\rho}\sum_{\hat{\vp}_i \in inlier}\|\hat{\vp}_i - \hat{\mathbf{d}}_i\|^2,
\end{equation}
where $\hat{\mathbf{d}}_i$ is the dominant direction of $\hat{\vp}_i$ that has been obtained previously, $\rho$ denotes the inlier ratio that is estimated in the outlier rejection step. We consider a scene with a small $\varepsilon$ as a Manhattan scene and perform the stitching process using the complete VPG scheme. Otherwise, we alternatively remove the VP guidance\footnote{A similar straighten scheme as in~\cite{brown2007automatic,chen2016natural} is adopted to make Eq.~\ref{eq:estimate_theta} solvable after removing the VP-relevant data term.} in Eq.~\ref{eq:estimate_theta} to make the proposed stitching algorithm fall back to the regular scheme as in~\cite{chen2016natural}.

\begin{figure*}[t]
	\begin{minipage}[t]{1.0\linewidth}
		\begin{minipage}[t]{0.49\columnwidth}
			\centering
			\captionsetup{font={small}}
			\begin{minipage}{0.49\linewidth}
				\includegraphics[width=1.0\linewidth]{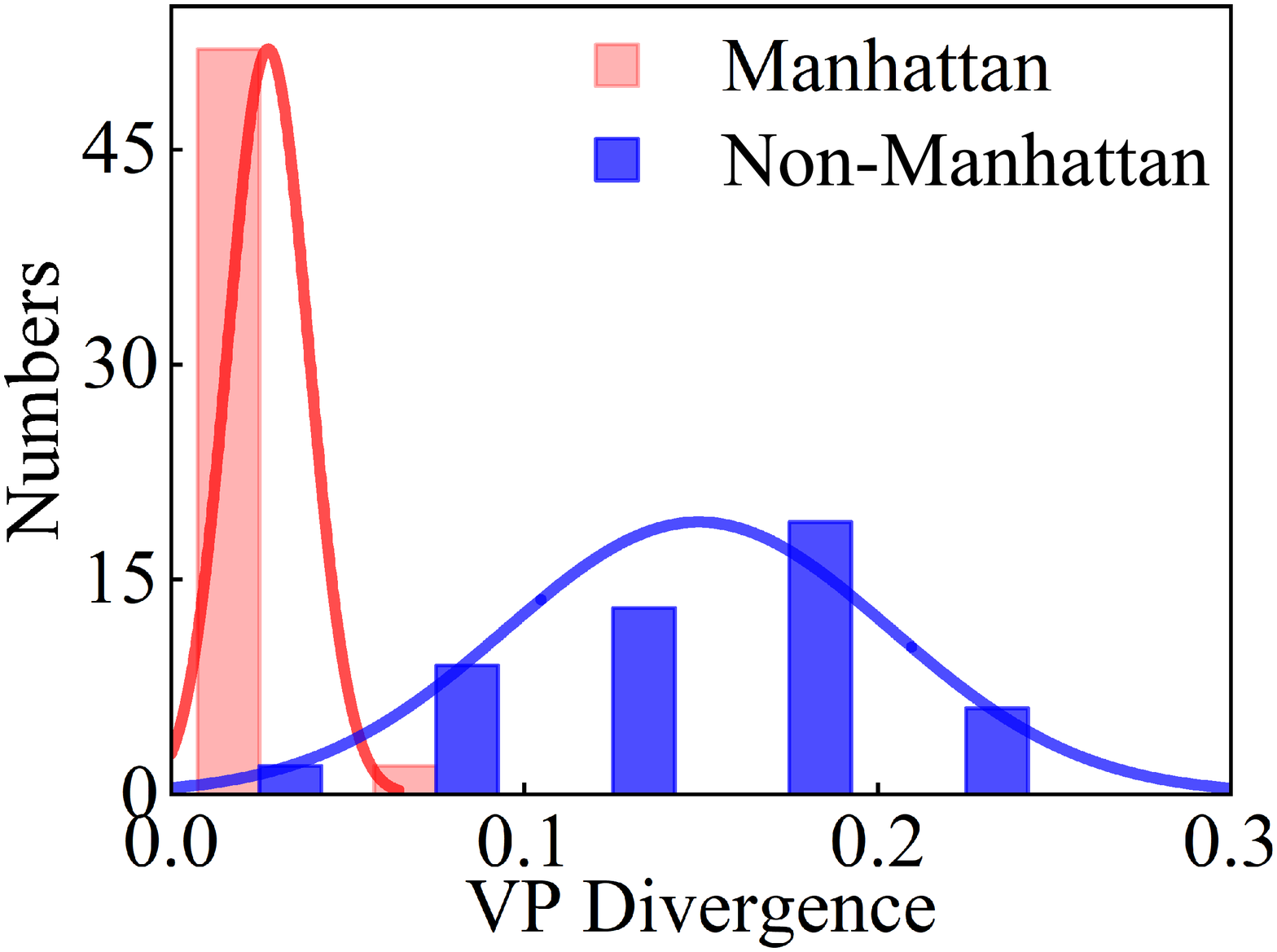}
				\centerline{(a)}
			\end{minipage}
			\hfill
			\begin{minipage}{0.49\linewidth}
				\includegraphics[width=1.0\linewidth]{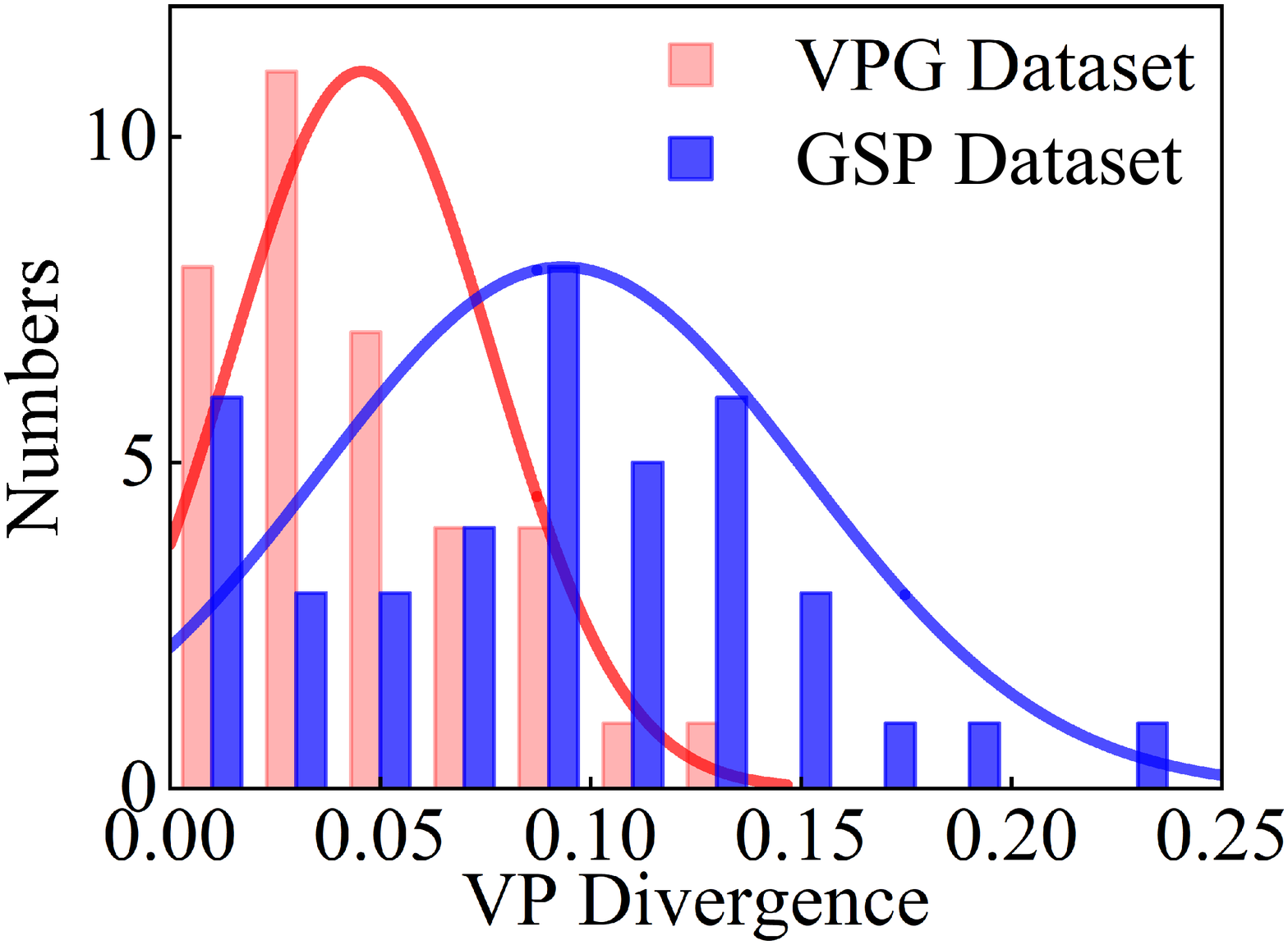}
				\centerline{(b)}
			\end{minipage}
			\captionof{figure}{(a) VP divergence distributions of Manhattan scenes and Non-Manhattan scenes. (b) The associated VP divergence distributions of the adopted VPG dataset and GSP dataset.}\vspace{-10pt}
			\label{fig:divergence}
		\end{minipage}
		\hfill
		\begin{minipage}[t]{0.49\columnwidth}
			\captionsetup{font={small}}
			\begin{minipage}[t]{1.0\columnwidth}
				\renewcommand{\tabcolsep}{4.0pt}
				\renewcommand\arraystretch{1.1}
				\captionof{table}{Average Runtime of different stitching methods.}
				\label{table:efficiency}
				\centerline{\begin{tabular}{|c|c|c|c|}
						\hline
						Dataset & AANAP & GSP & VPG \\\hline
						VPG Dataset & 109.44s & 45.82s & 47.82s \\\hline
						GSP Dataset & 53.63s & 23.55s & 29.04s \\\hline
				\end{tabular}}\vspace{-18pt}
			\end{minipage}
		\end{minipage}
	\end{minipage}
	\vfill
	\vspace{12pt}
	\begin{minipage}[t]{1.0\linewidth}
		\centering
		\captionsetup{font={small}}
		\begin{minipage}{0.245\linewidth}
			\includegraphics[width=1.0\linewidth]{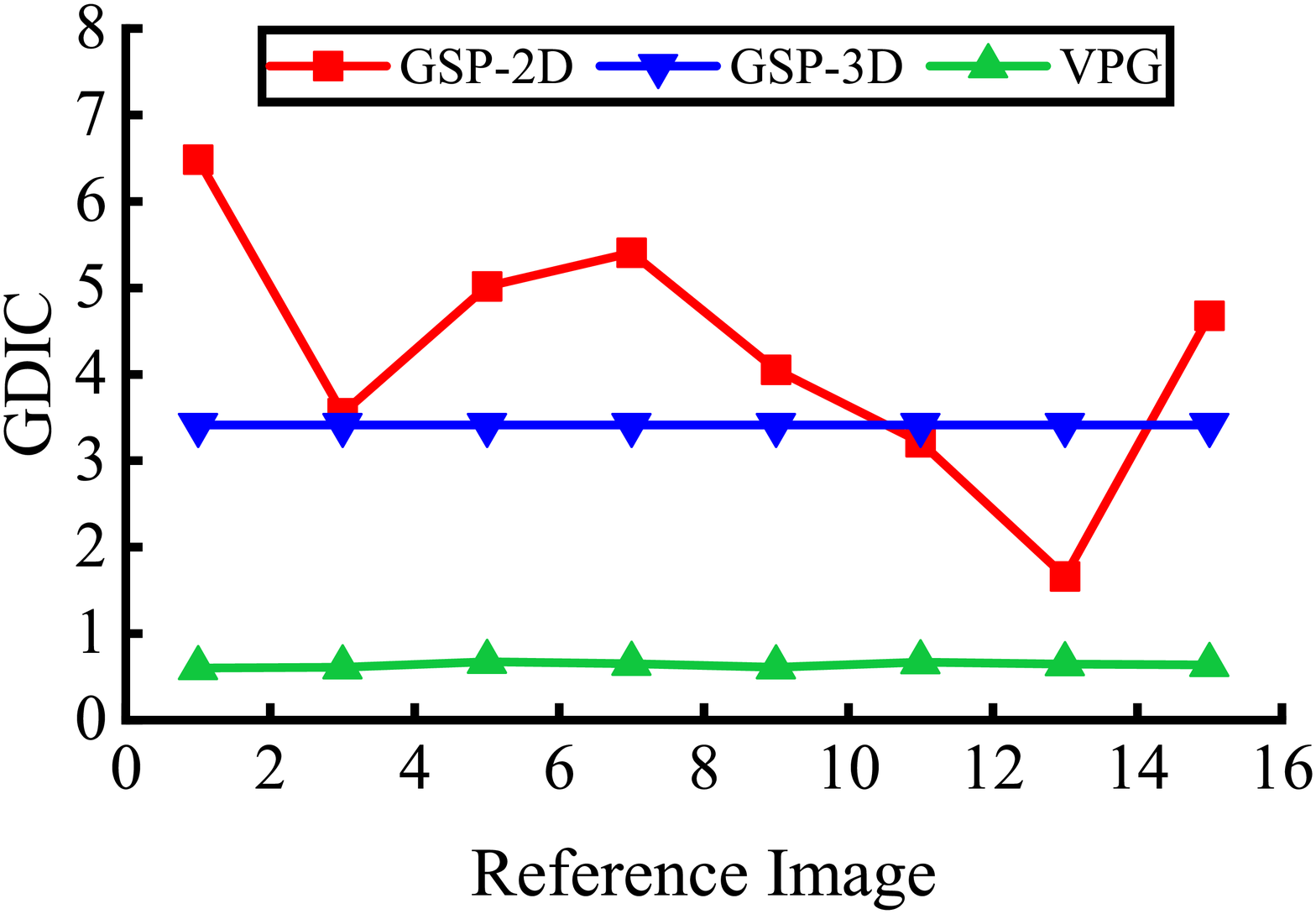}
			\centerline{(a)}
		\end{minipage}
		\hfill
		\begin{minipage}{0.245\linewidth}
			\includegraphics[width=1.0\linewidth]{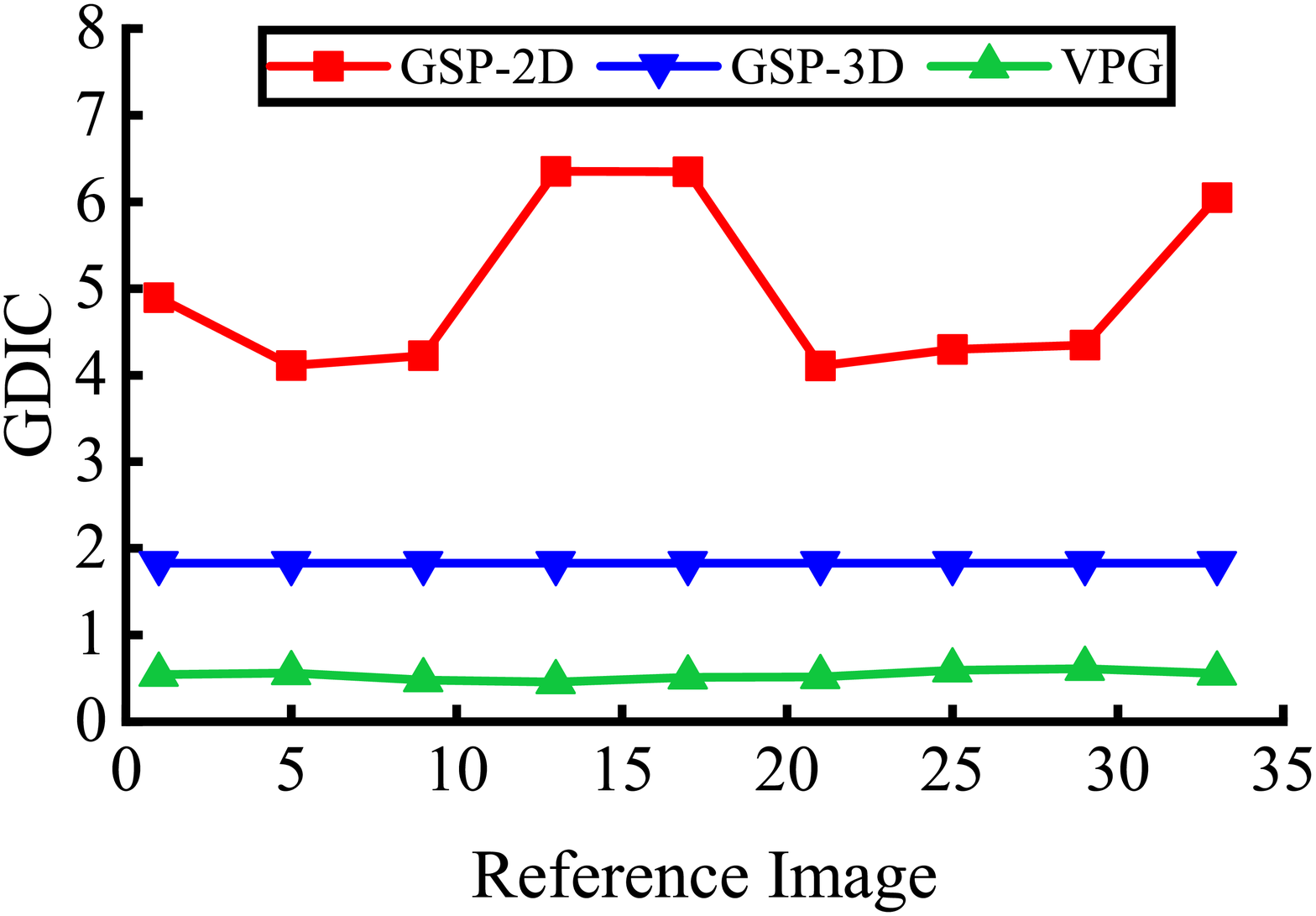}
			\centerline{(b)}
		\end{minipage}
		\hfill
		\begin{minipage}{0.245\linewidth}
			\includegraphics[width=1.0\linewidth]{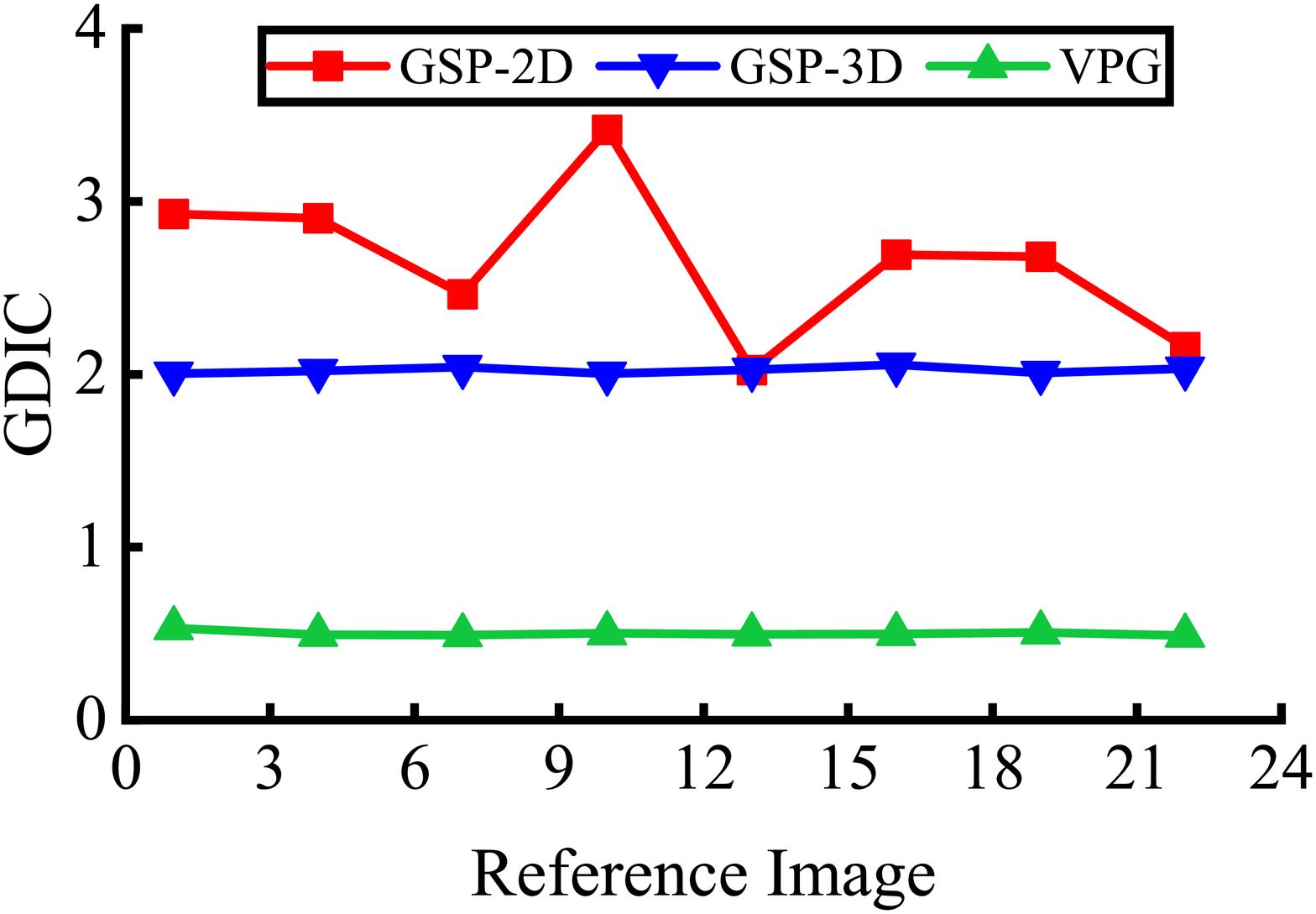}
			\centerline{(c)}
		\end{minipage}
		\hfill
		\begin{minipage}{0.245\linewidth}
			\includegraphics[width=1.0\linewidth]{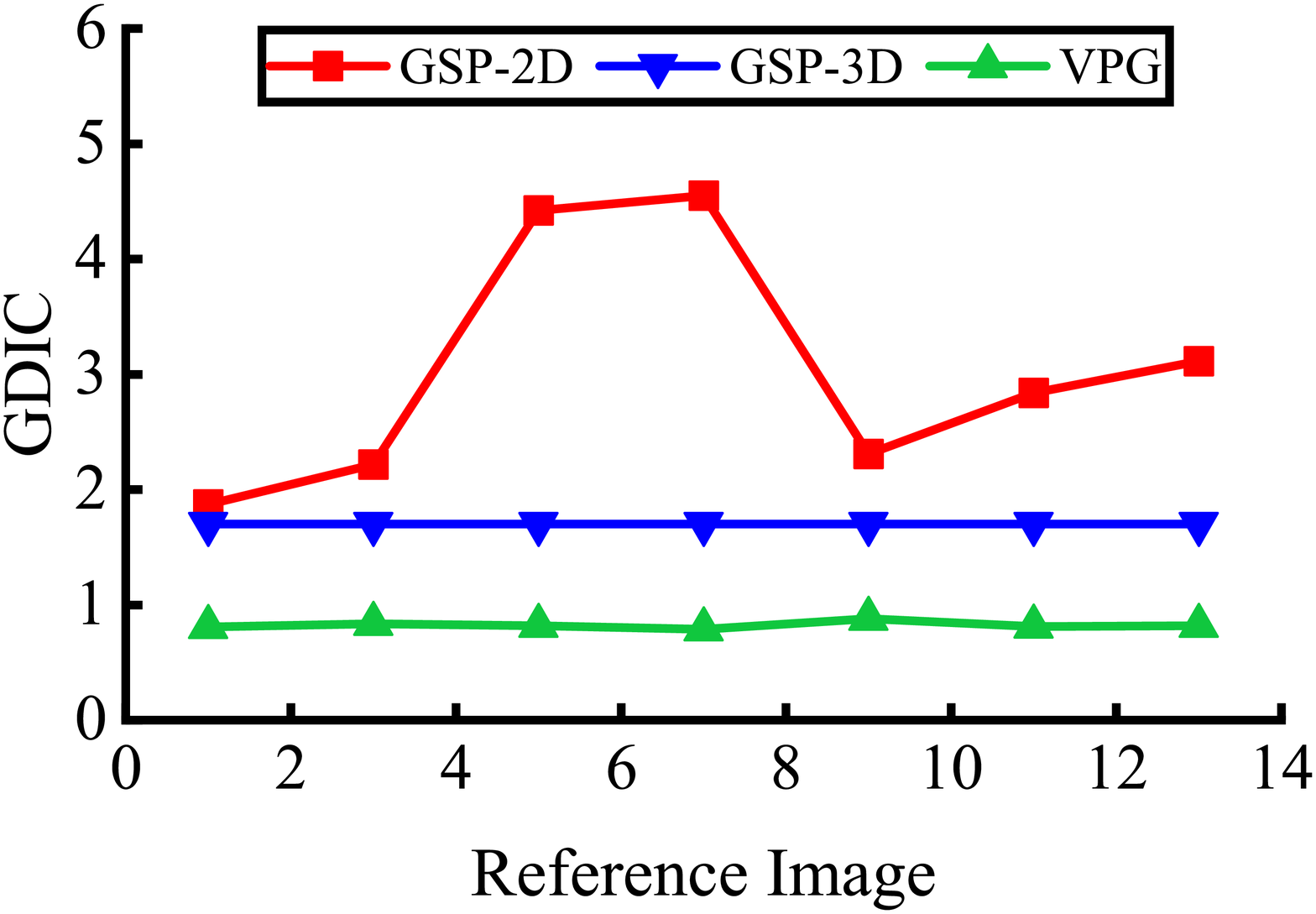}
			\centerline{(d)}
		\end{minipage}
		\captionof{figure}{Quantitative evaluations of the influence of reference selection on different stitching methods. (a)-(d) present the results on VPG-$03$, $05$, $08$ and $11$ respectively.}\vspace{-18pt}
		\label{fig:reference_selection}
	\end{minipage}	
\end{figure*}

In order to determine a suitable threshold for $\varepsilon$, a similar statistical analysis scheme as in~\cite{liu2016meshflow} is applied. Specifically, we first collected another $130$ image sets~(excluding the above data for algorithm evaluation). Half of them were captured in Manhattan scenes, and another half of them were captured in natural scenes~(non-Manhattan). Their VP divergence values were computed according to Eq.~\ref{eq:vp_divergence} and are presented in Figure~\ref{fig:divergence}. As we can see, $\varepsilon$ from a Manhattan scene usually is small and tends to be limited in a narrow range. On the contrary, $\varepsilon$ of a natural scene usually has a relatively large value. $\varepsilon_0=0.10$ seems to be a valid indicator threshold to distinguish the Manhattan scenes from non-Manhattan scenes.

\begin{figure*}[t]
	\centering
	\captionsetup{font={small}}
	\begin{minipage}{1.0\linewidth}
		\includegraphics[width=1.0\linewidth]{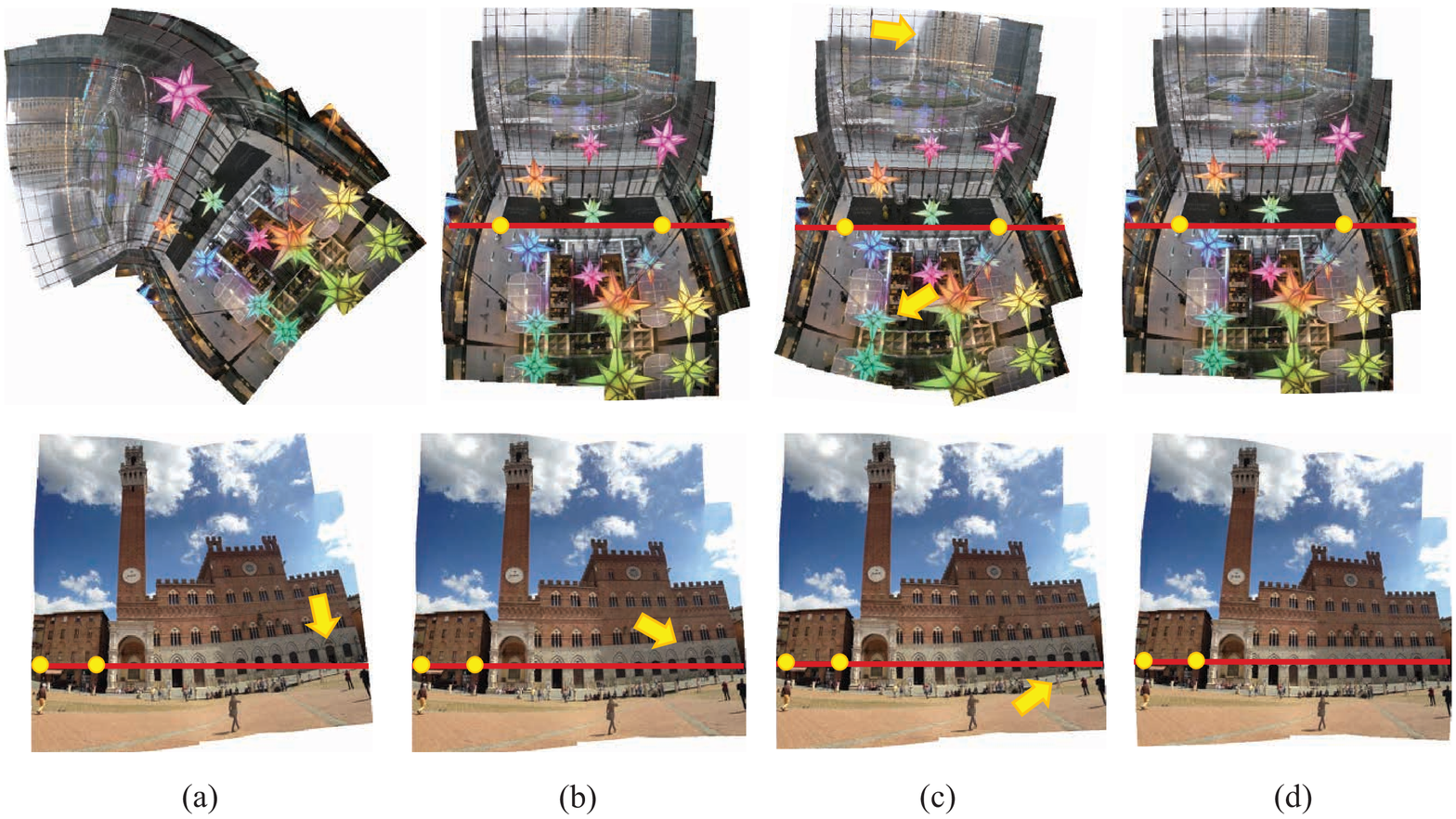}
		\captionof{figure}{Qualitative comparisons with AANAP~\cite{lin2015adaptive} and GSP~\cite{chen2016natural} on $2$ GSP sets. (a) results from AANAP. (b) results from GSP-2D. (c) results from GSP-3D. (d) results from VPG. Top row is a indoor scene with $35$ input images and the associate $\varepsilon=0.044\leq\varepsilon_0$. Bottom row is a outdoor scene with $5$ input images and $\varepsilon=0.048\leq\varepsilon_0$.}
		\label{fig:comparison_gsp_manhattan}
	\end{minipage}
	\vfill
	\begin{minipage}{1.0\linewidth}
		\centering
		\begin{minipage}{0.85\linewidth}
			\begin{minipage}{0.05\linewidth}
				\text{(a)}
			\end{minipage}
			\hfill
			\begin{minipage}{0.95\linewidth}
				\includegraphics[width=1.0\linewidth]{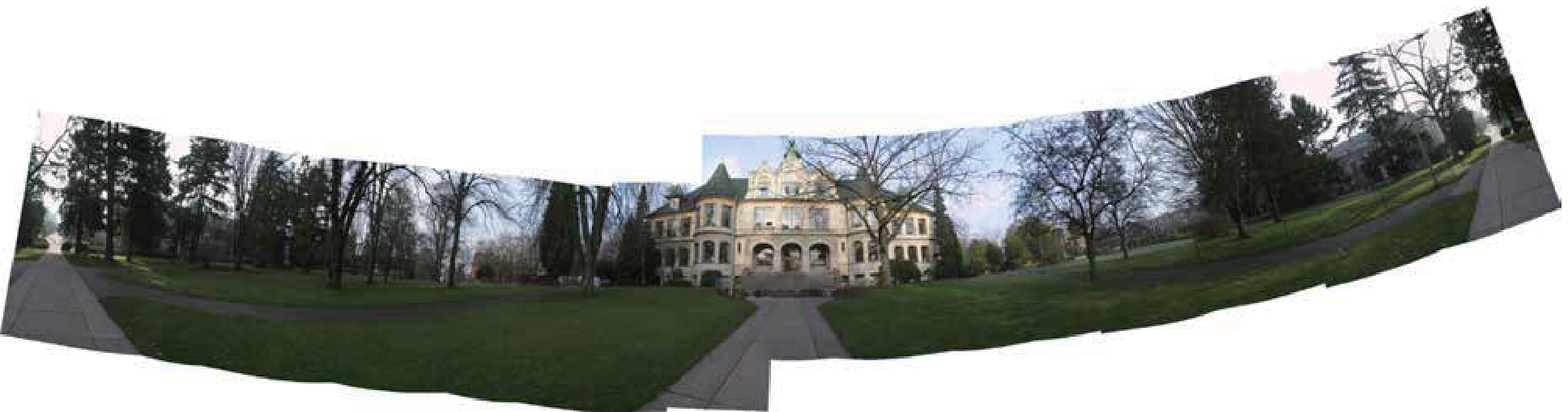}
			\end{minipage}
		\end{minipage}
		\vfill
		\begin{minipage}{0.85\linewidth}
			\begin{minipage}{0.05\linewidth}
				\text{(b)}
			\end{minipage}
			\hfill
			\begin{minipage}{0.95\linewidth}
				\includegraphics[width=1.0\linewidth]{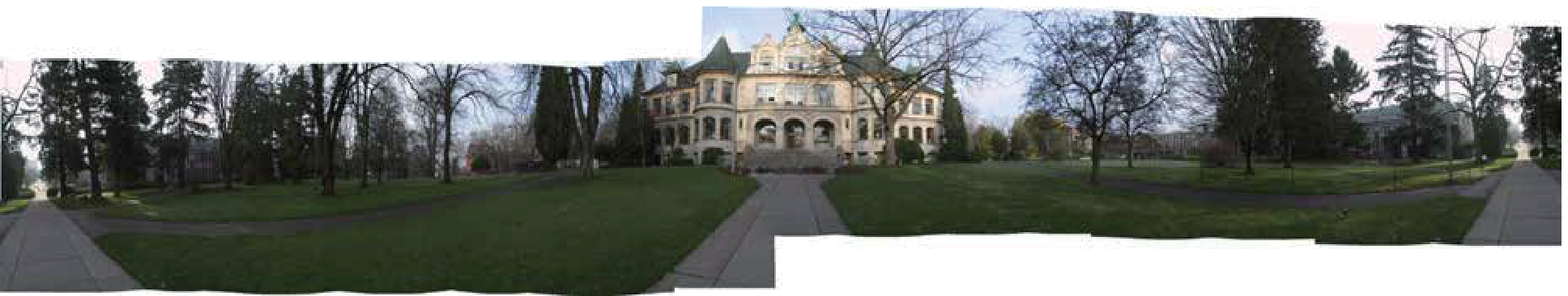}
			\end{minipage}
		\end{minipage}
		\vfill
		\begin{minipage}{0.85\linewidth}
			\begin{minipage}{0.05\linewidth}
				\text{(c)}
			\end{minipage}
			\hfill
			\begin{minipage}{0.95\linewidth}
				\includegraphics[width=1.0\linewidth]{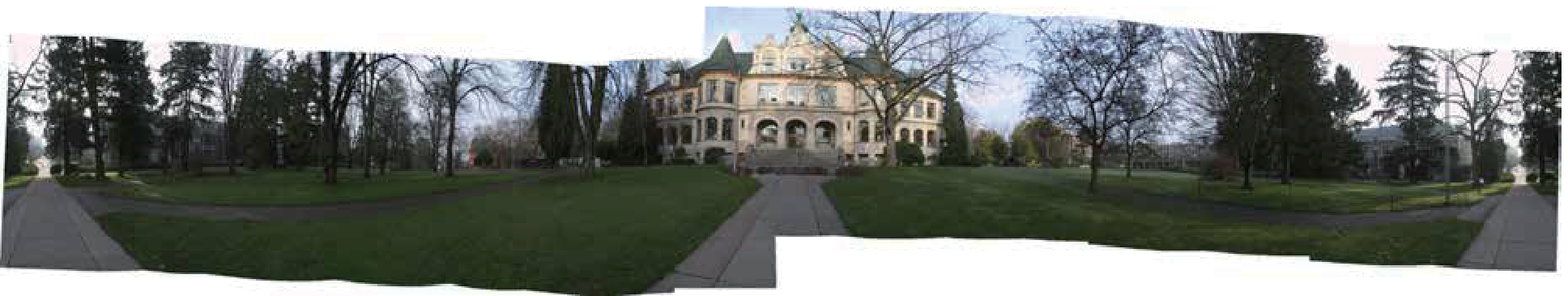}
			\end{minipage}
		\end{minipage}
		\vfill
		\begin{minipage}{0.85\linewidth}
			\begin{minipage}{0.05\linewidth}
				\text{(d)}
			\end{minipage}
			\hfill
			\begin{minipage}{0.95\linewidth}
				\includegraphics[width=1.0\linewidth]{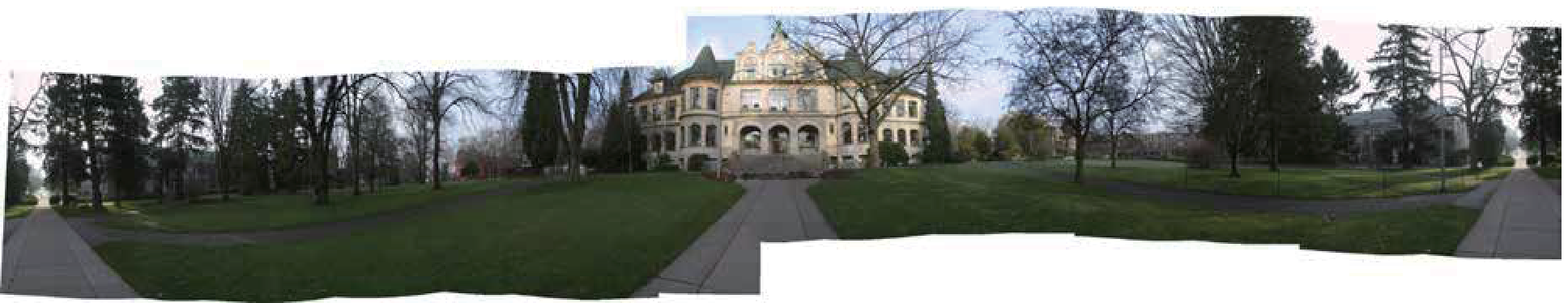}
			\end{minipage}
		\end{minipage}
		\captionof{figure}{Qualitative comparisons with AANAP~\cite{lin2015adaptive} and GSP~\cite{chen2016natural} on one GSP set. (a) results from AANAP. (b) results from GSP-2D. (c) results from GSP-3D. (d) results from our VPG. It is a scene with $15$ input images and $\varepsilon=0.161>\varepsilon_0$.}
		\label{fig:comparison_gsp_non_manhattan_1}
	\end{minipage}
\end{figure*}

\begin{figure*}[t]
	\centering
	\captionsetup{font={small}}
	\begin{minipage}{1.0\linewidth}
		\centering
		\renewcommand{\tabcolsep}{3.5pt}
		\renewcommand\arraystretch{1.1}
		\captionof{table}{Basic information about the $12$ panoramas (a)-(l) presented in Figure~\ref{fig:gsp_gallery}. Note that the $\varepsilon$ values from (a)-(h) are less than $\varepsilon_0$ while values from (i)-(l) are larger than $\varepsilon_0$.}
		\label{table:gsp_description}
		\centering
		\normalsize
		\begin{tabular}{|c|c|c|c|c|c|c|c|c|c|c|c|c|}
			\hline
			 No. & (a) & (b) & (c) & (d) & (e) & (f) & (g) & (h) & (i) & (j) & (k) & (l)\\\hline
			 Image Numbers & 2 & 4 & 5 & 6 & 11 & 3 & 5 & 21 & 10 & 5 & 7 & 15\\\hline
			 $\varepsilon$ & 0.0204 & 0.0019 & 0.0068 & 0.0005 & 0.0740 & 0.0002 & 0.0114 & 0.0781 & 0.1917 & 0.1291 & 0.1464 & 0.2022\\\hline
		\end{tabular}
	\end{minipage}
	\begin{minipage}{1.0\linewidth}
		\centering
		\includegraphics[width=0.9\linewidth]{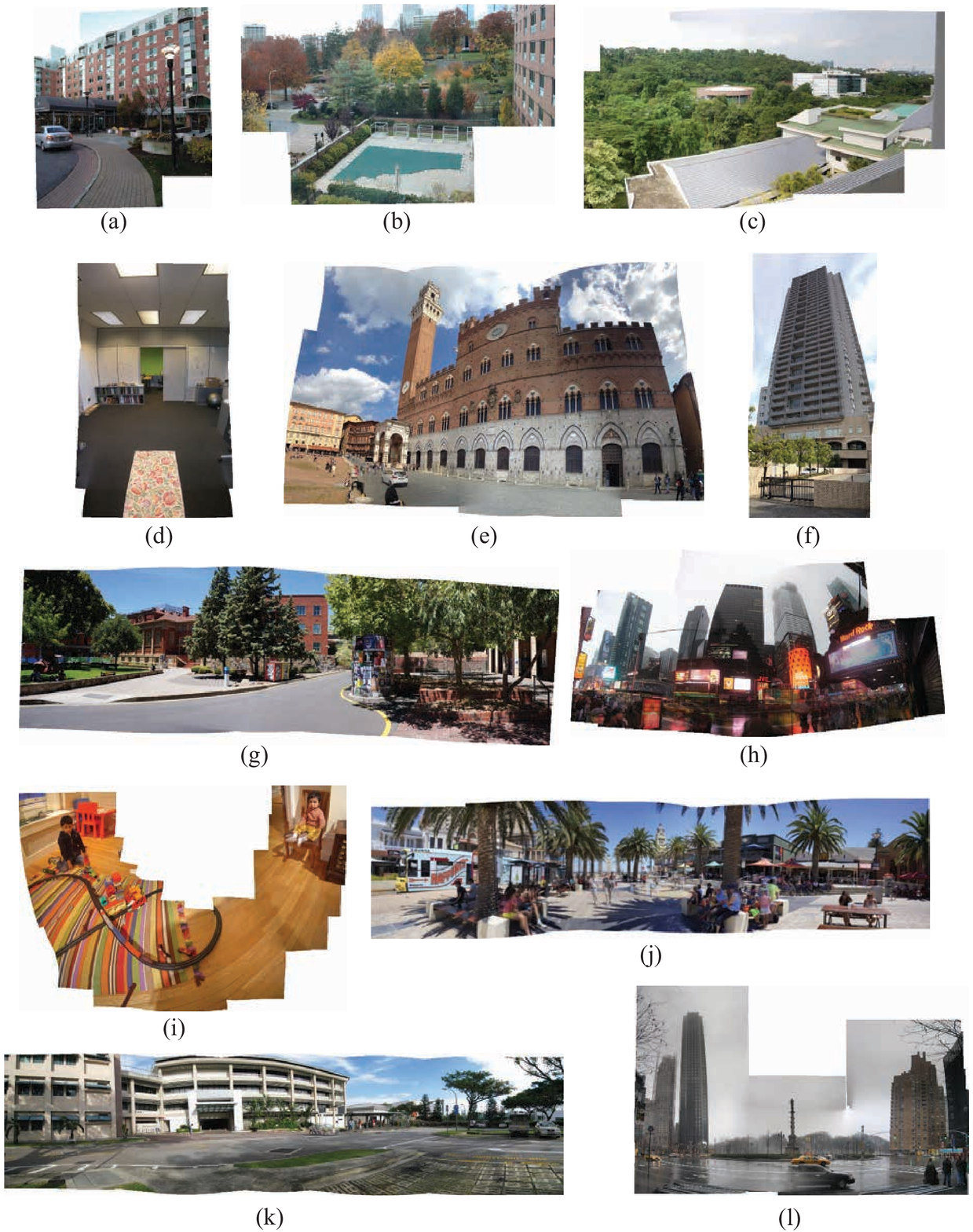}
		\captionof{figure}{More stitching results on GSP dataset.}
		\label{fig:gsp_gallery}
	\end{minipage}
\end{figure*}

To verify the adaptability of VPG when resorting to $\varepsilon$, we hope to simulate a practical VPG application scenario in which the regularity prior about the scene~(Manhattan or non-Manhattan) is unknown and so we tested VPG on the GSP dataset~\cite{chen2016natural}. It consists of $42$ image sets and contains nearly all popular images for stitching algorithm evaluation. Since the Manhattan assumption is not necessarily satisfied in GSP dataset, as presented in Figure~\ref{fig:divergence}, the associated VP divergences are distributed on both sides of $\varepsilon_0$. Figure~\ref{fig:comparison_gsp_manhattan} and Figure~\ref{fig:comparison_gsp_non_manhattan_1} present typical results on GSP dataset. On the one hand, when the $\varepsilon$ of a scene is small, VPG manages to improve the panorama naturalness by utilizing the reliable VPs guidance. On the other hand, if the $\varepsilon$ is large which indicates the scene is prone to be a non-Manhattan scene, VPG still can produce natural looking panoramas by weakening the extracted VPs guidance. Note that all results are produced by VPG automatically without any manual intervention. It demonstrates that the proposed VPG can be well applied in general~(Manhattan or non-Manhattan) scenes. Figure~\ref{fig:gsp_gallery} presents more stitching results on GSP dataset.

\subsection{Stability for Reference Selection}\label{subsec:analysis_reference}\vspace{-3pt}
Reference selection is an important but challenging issue for image stitching. Many algorithms are sensitive to this step and as a result, they may yield significantly different panoramas when different images are selected as the reference. Selecting the optimal reference is not easy even many methods have been proposed in the past decade~\cite{choe2006optimal,szeliski2007image, xie2018robust,xia2017globally}. Taking $4$ sets of synthetic images as an example, Figure~\ref{fig:reference_selection} reports the GDIC quantitative results of VPG when different images are selected as the reference during the stitching process. As we can see, results from GSP-2D are severely affected by reference selection since different reference will lead to different images with the zero-rotation, which makes the panorama appearance change significantly. In contrast, the proposed VPG produces much more stable GDIC values no matter which image is chosen as the reference. Note that results from GSP-3D have a similar stability with VPG, but the corresponding GDIC values are much larger which indicate a worse panorama naturalness.\vspace{-7pt}

\subsection{Scalability for Higher Alignment Accuracy}\label{subsec:analysis_scalability}
\begin{figure*}[t]
	\centering
	\captionsetup{font={small}}
	\begin{minipage}{0.95\linewidth}
		\includegraphics[width=1.0\linewidth]{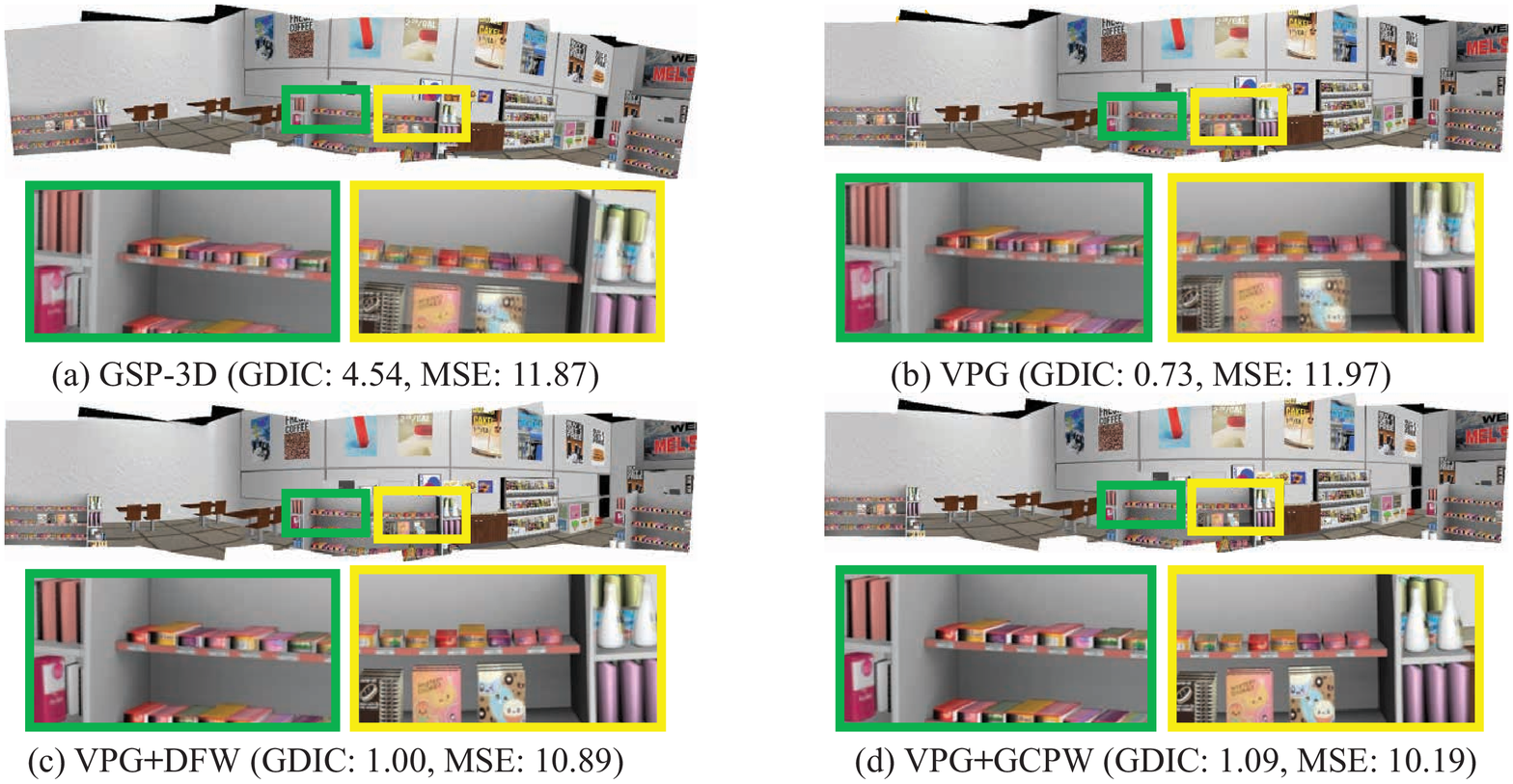}
		\captionof{figure}{Qualitative comparisons between GSP-3D~\cite{chen2016natural}, the original VPG, VPG+DFW and VPG+GCPW on VPG synthetic set $01$. By combining VPG with DFW and GCPW, the alignment accuracy gets improved significantly without any obvious loss of panorama naturalness.}
		\label{fig:alignment01}
	\end{minipage}
	\vfill
	\begin{minipage}{0.95\linewidth}
		\includegraphics[width=1.0\linewidth]{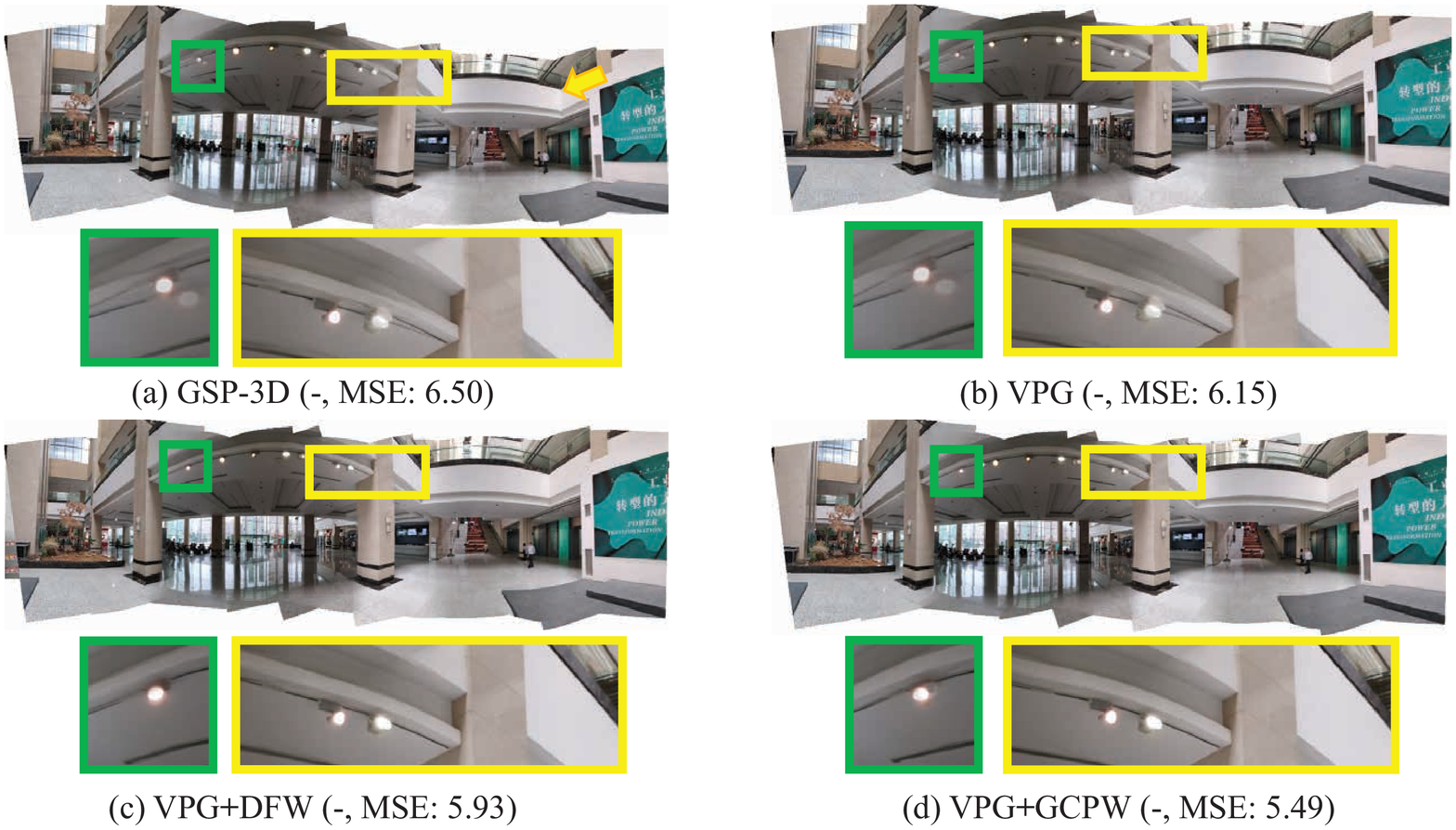}
		\captionof{figure}{Qualitative comparisons between GSP-3D~\cite{chen2016natural}, the original VPG, VPG+DFW and VPG+GCPW on VPG real set $13$. By combining VPG with DFW and GCPW, the alignment accuracy gets improved significantly without any obvious loss of panorama naturalness. Note that the GDIC values are unavilable for real image data.}
		\label{fig:alignment17}
	\end{minipage}
\end{figure*}

\vspace{-3pt}
Except for naturalness, alignment accuracy is another essential issue that is widely considered when designing a stitching algorithm. Some methods~\cite{lin2011smoothly} achieve high panorama naturalness at the expense of a decreased alignment accuracy. In previous experiments and analysis, for a fair comparison, VPG follows the same scheme as in~\cite{chen2016natural} to extract the alignment constraints. Although it inherits the powerful alignment capability provided by APAP~\cite{zaragoza2013projective}, the alignment accuracy can be further increased by many recent advanced stitching frameworks like DFW~\cite{li2015dual} and GCPW~\cite{chen2018generalized}. In this section, we show that the proposed VPG is scalable for achieving a higher alignment accuracy. In other words, the naturalness improvement achieved by our VPG is compatible with high alignment accuracy. Figure~\ref{fig:alignment01} and Figure~\ref{fig:alignment17} present two groups of panoramas produced by different methods and report the associated GDIC values and the MSE of alignment accuracy~\cite{chen2018multiple,chen2018video,lin2016seagull,lin2017direct}. By combining VPG with DFW and GCPW, the output panoramas not only have more natural looks than GSP, but also have higher alignment accuracy than both GSP and the original VPG.\vspace{-5pt}

\subsection{Time Efficiency}\label{subsec:analysis_efficiency}\vspace{-3pt}
To evaluate the stitching efficiency of VPG, Table~\ref{table:efficiency} reports the average runtime of AANAP, GSP, and VPG on 36-set VPG dataset and 42-set GSP dataset respectively. Without any acceleration technology, the proposed VPG method slightly increases the runtime by about $10\%$ when compared with GSP but it significantly improves the panorama naturalness. Besides, VPG is approximately $2$ times faster than AANAP with a much better panorama naturalness.\vspace{-11pt}

\section{Conclusions and Future Work}\vspace{-5pt}
In this paper, we propose a vanishing-point-guided stitching method called VPG. VPG successfully exploits the predominance of VPs to achieve a robust estimation for image similarity prior, which finally leads to a more natural looking panorama. Quantitative and qualitative comparisons on synthetic and real images combined with a user study demonstrate VPG's superiority over other state-of-the-art methods. More analyses upon VPG show that although VPG is designed for Manhattan scenes, it possesses good adaptability for general scenes through a degradation mechanism. Meanwhile, due to the introduction of global VPs, VPG outputs stable panoramas that are free from different reference selections. Moreover, VPG is scalable and compatible with other advanced stitching frameworks to achieve a coordination between panorama naturalness and alignment accuracy. We also observed some limitations for VPG. Firstly, VPG is prone to fall back to the non-Manhattan scene when the number of involved images is small~(\eg, $2$ or $3$). Because in such cases, the VP consistency usually is not remarkable for similarity prior estimation. Secondly, VPG may fail when facing dramatic depth variation or large parallax, which could influence the VPs alignment results and cause unnatural artifacts. Note that extremely large parallax is also challenging for most other existing stitching approaches and is the issue that needs to be overcome in the future work.

\ifCLASSOPTIONcaptionsoff
  \newpage
\fi



%

\clearpage
\bibliography{egbib}
%








\end{document}